\documentclass[runningheads]{llncs}
\usepackage[T1]{fontenc}
\usepackage{graphicx}
\usepackage{booktabs} 
\usepackage[misc]{ifsym}
\newcommand{\corr}{(\Letter)}
\usepackage{cite}
\usepackage{mwe}
\usepackage{hyperref}       
\usepackage{url}            
\usepackage{amsfonts}       
\usepackage{mathtools}      
\usepackage{array}
\usepackage{nicefrac}       
\usepackage{tikz}
\usetikzlibrary{arrows.meta, matrix, calc, decorations.pathreplacing,
                ext.paths.ortho, positioning, quotes}                
\usetikzlibrary{positioning,fit,backgrounds,shadows}
\usetikzlibrary{shapes.multipart, decorations.pathmorphing}

\usepackage{rotating} 

\usepackage{algorithm} 
\usepackage{algorithmic}  
\usepackage[algo2e]{algorithm2e} 

\renewcommand{\arraystretch}{1.4}
\definecolor{softblue}{RGB}{200, 220, 255}
\definecolor{softorange}{RGB}{255, 220, 180}
\definecolor{softred}{RGB}{255, 230, 230}
\definecolor{softgrey}{RGB}{245, 245, 245}

\newcommand{\coloredblockscontig}[3]{
    \foreach \i in {0,...,#2} {
        \begin{scope}[shift={(#1+\i,0,0)}]
            \draw[fill=#3, draw=black] (0,0,0) -- ++(0,1,0) -- ++(0,0,1) -- ++(0,-1,0) -- cycle;
            \draw[fill=#3!80, draw=black] (0,0,1) -- ++(0,1,0) -- ++(1,0,0) -- ++(0,-1,0) -- cycle;
            \draw[fill=#3!60, draw=black] (0,1,0) -- ++(1,0,0) -- ++(0,0,1) -- ++(-1,0,0) -- cycle;
            \draw[fill=#3!40, draw=black] (1,0,0) -- ++(0,1,0) -- ++(0,0,1) -- ++(0,-1,0) -- cycle;
        \end{scope}
    }
}

\begin{document}

\title{Large Causal Models for Temporal Causal Discovery}
\titlerunning{Large Causal Models for Temporal Causal Discovery}

\author{Nikolaos Kougioulis\inst{1,2} \corr \and
Nikolaos Gkorgkolis\inst{1,2} \and
MingXue Wang\inst{3} \and
Bora Caglayan\inst{3} \and
Dario Simionato\inst{3} \and
Andrea Tonon\inst{3} \and
Ioannis Tsamardinos\inst{1,2}}

\toctitle{Large Causal Models for Temporal Causal Discovery}
\tocauthor{Nikolaos Kougioulis, Nikolaos Gkorgkolis, MingXue Wang, Bora Caglayan, Dario Simionato, Andrea Tonon, Ioannis Tsamardinos}

\authorrunning{N. Kougioulis et al.}

\institute{
Institute of Applied \& Computational Mathematics, FORTH \and
Computer Science Department, University of Crete \\
\email{nikolaos.kougioulis@iacm.forth.gr} \\
\email{\{gkorgkolis, tsamard\}@csd.uoc.gr}\\
\and
Huawei Ireland Research Centre, Dublin, Ireland \\
\email{\{wangmingxue1, bora.caglayan\}@huawei.com}\\
\email{dario.simionato1@h-partners.com}\\
\email{andrea.tonon1@huawei-partners.com}
}

\maketitle

\begin{abstract}
Causal discovery for both cross-sectional and temporal data has traditionally followed a dataset-specific paradigm, where a new model is fitted for each individual dataset. Such an approach limits the potential of multi-dataset pretraining.  The concept of \textit{large causal models (LCMs)} envisions a class of pre-trained neural architectures specifically designed for temporal causal discovery. Prior approaches are constrained to small variable counts, degrade with larger inputs, and rely heavily on synthetic data, limiting generalization. We propose a principled framework for LCMs, combining diverse synthetic generators with realistic time-series datasets, allowing learning at scale. Extensive experiments on synthetic, semi-synthetic and realistic benchmarks show that LCMs scale effectively to higher variable counts and deeper architectures while maintaining strong performance. Trained models achieve competitive or superior accuracy compared to classical and neural baselines, particularly in out-of-distribution settings, while enabling fast, single-pass inference. Results demonstrate LCMs as a promising foundation-model paradigm for temporal causal discovery. Experiments and model weights are available at \href{https://github.com/kougioulis/LCM}{\texttt{https://github.com/kougioulis/LCM}}.
\end{abstract}


\section{Introduction} \label{sec:introduction}

Causal discovery (CD) aims to recover the causal structure of data \cite{spirtes2001causation, pearl2009causality}. In many domains, data are inherently temporal, motivating causal discovery from multivariate time series \cite{runge2018causal, runge2019inferring, runge2023causal}.

Classical temporal CD methods provide strong identifiability guarantees under explicit assumptions \cite{spirtes2001causation, runge2018causal}, yet rely on conditional independence testing or combinatorial search, whose computational and statistical complexity scales poorly with input dimensionality. Methods assuming known functional forms (e.g., linearity) \cite{pamfil2020dynotears, hyvarinen2010estimation} degrade when assumptions are violated.

Neural methods have been introduced that attempt to amortize CD, include variational approaches \cite{lorch2022amortized, ashman2023causal, geffner2024deep}, autoregressive and attention-based models \cite{ke2023learning}, and adversarial-based frameworks \cite{goudet2017causal, goudet2018learning}. While promising, most remain dataset-specific and struggle to generalize across heterogeneous systems, especially in zero-shot settings.

In parallel, foundation models (FMs) \cite{bommasani2021opportunities} have demonstrated that large neural architectures trained on diverse data can generalize effectively across tasks \cite{radford2018improving, kolesnikov2021an, liu2024atimer, liu2024btimer, das2024decoder, wu2025transformers}, thus motivating \textit{Large Causal Models (LCMs)}: pre-trained FMs that can leverage multiple datasets and perform efficient, fast temporal CD. Prior work \cite{stein2024embracing} remains proof-of-concept, limited to low dimensions and narrow synthetic training distributions.

Importantly, a key obstacle to scaling large causal models is the scarcity of large, diverse time-series datasets with ground-truth causal graphs, with prior work relying on synthetic generators \cite{assaad2022survey}, which limits generalization.

In this work, we demonstrate that LCMs can be made scalable and robust through data-centric
training on a large and heterogeneous corpus, enabling reliable zero-shot generalization and systematic evaluation across diverse temporal settings. Our contributions can be summarized as follows:

\begin{enumerate}
\item We formalize \textit{Large Causal Models (LCMs)} as FMs for temporal CD across heterogeneous data distributions, replacing dataset-specific algorithms with foundation model-style pretraining,
\item We demonstrate that scaling failures in prior work stem from insufficient training distribution diversity and propose the inclusion of realistic data, 
\item We curate a \textit{large-scale collection} of time-series samples and causal graph pairs suitable for both training of LCMs and benchmarking CD methods,
\item We demonstrate superior or competitive performance compared to classical CD methods, with substantially reduced runtime via single-pass CD.
\end{enumerate}

\paragraph{Temporal Structural Causal Models (TSCMs).} A Structural Causal Model (SCM) represents each endogenous variable as a function of its causal parents and an exogenous noise term \cite{spirtes2001causation, pearl2009causality, peters2017elements}. For a multivariate time series \(\mathbf{V}_t = (V^1_t, \ldots, V^k_t)'\), a Temporal SCM (TSCM) is defined as \cite{runge2018causal} \(V^j_t \coloneqq f^j\!\left(\text{Pa}(V^j_t)\right) + \epsilon^j_t, \quad t \in \mathbb{Z}\), where \(\text{Pa}(V^j_t)\) denotes the set of lagged causal parents drawn from the past \(\ell_{\max}\) time steps, \(\text{Pa}(V^j_t) \subseteq \{\mathbf{V}_{t-1}, \ldots, \mathbf{V}_{t-\ell_{\max}}\}\). Here, \(\ell_{\max}\) is a fixed \textit{maximum causal lag}. We assume additive noise models, causal stationarity, no latent confounders, and no contemporaneous effects (i.e. effects within a lag \(\ell=0\)). Detailed definitions are provided in Appendix \ref{app:definitions}.

\paragraph{Lagged Causal Graphs.} Temporal causal relationships in a TSCM are represented by a lagged causal graph (window graph) \cite{assaad2022survey}, a DAG whose nodes correspond to time-indexed variables and whose directed edges are of the form \(V^i_{t-\ell} \rightarrow V^j_t\) for \(\ell \in \{1,\ldots,\ell_{\max}\}\). Edges are oriented forward in time and denote direct causal effects, while indirect effects are encoded via directed paths.

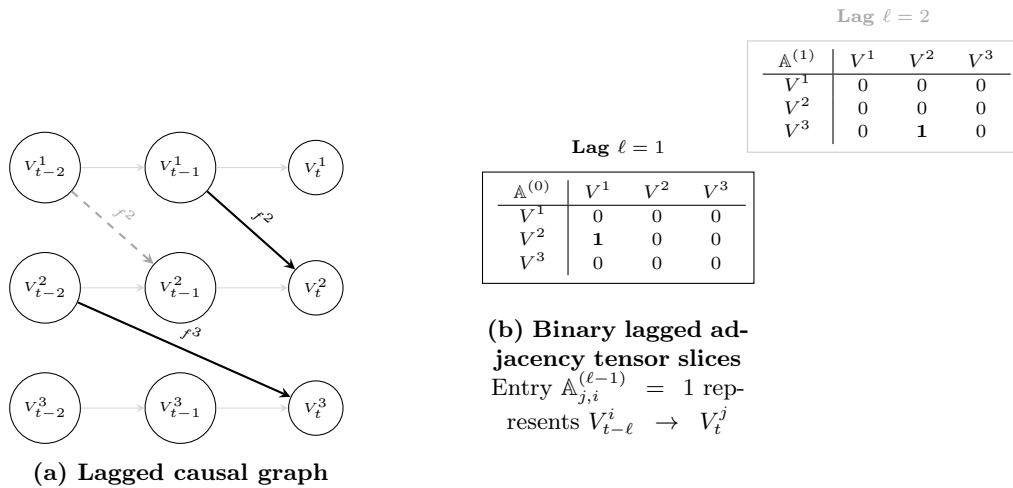
\begin{figure}[t]
\centering
\setlength{\abovecaptionskip}{4pt}
\setlength{\belowcaptionskip}{-4pt}

\begin{minipage}[t]{0.43\textwidth}
\centering
\begin{tikzpicture}[
    varnode/.style={circle, draw, minimum size=0.55cm, font=\tiny},
    >=stealth,
    ->,
    dashededge/.style={->, dashed, thick, color=gray!70},
    solidedge/.style={->, thick},
]

\node[varnode] (V1t2) at (0,3.2) {$V^1_{t-2}$};
\node[varnode] (V2t2) at (0,1.6) {$V^2_{t-2}$};
\node[varnode] (V3t2) at (0,0) {$V^3_{t-2}$};

\node[varnode] (V1t1) at (1.8,3.2) {$V^1_{t-1}$};
\node[varnode] (V2t1) at (1.8,1.6) {$V^2_{t-1}$};
\node[varnode] (V3t1) at (1.8,0) {$V^3_{t-1}$};

\node[varnode] (V1t) at (3.6,3.2) {$V^1_{t}$};
\node[varnode] (V2t) at (3.6,1.6) {$V^2_{t}$};
\node[varnode] (V3t) at (3.6,0) {$V^3_{t}$};

\foreach \i in {1,2,3} {
  \draw[->, gray!30] (V\i t2) -- (V\i t1);
  \draw[->, gray!30] (V\i t1) -- (V\i t);
}

\draw[solidedge] (V1t1) -- (V2t)
  node[midway, above, sloped, font=\tiny] {$f^{2}$};

\draw[dashededge] (V1t2) -- (V2t1)
  node[midway, above, sloped, font=\tiny] {$f^{2}$};

\draw[solidedge] (V2t2) -- (V3t)
  node[midway, above, sloped, font=\tiny] {$f^{3}$};

\end{tikzpicture}
\vspace{0.3em}\\
{\small \textbf{(a) Lagged causal graph}}
\end{minipage}
\hspace{0.01\textwidth}
\begin{minipage}[t]{0.53\textwidth}
\centering
\begin{tikzpicture}

\node (tab2) [draw=gray!40, fill=white] {
    \scriptsize
    \renewcommand{\arraystretch}{1.05}
    \scalebox{0.7}{\begin{tabular}{c|ccc}
        $\mathbb{A}^{(1)}$ & $V^1$ & $V^2$ & $V^3$ \\ \hline
        $V^1$ & 0 & 0 & 0 \\
        $V^2$ & 0 & 0 & 0 \\
        $V^3$ & 0 & \textbf{1} & 0 \\
    \end{tabular}}
};
\node[above=0.08cm of tab2, font=\scriptsize\bfseries, text=gray!70] {Lag $\ell=2$};

\node (tab1) [draw=black, fill=white, below left=0.25cm and -0.1cm of tab2] {
    \scriptsize
    \renewcommand{\arraystretch}{1.05}
    \scalebox{0.7}{\begin{tabular}{c|ccc}
        $\mathbb{A}^{(0)}$ & $V^1$ & $V^2$ & $V^3$ \\ \hline
        $V^1$ & 0 & 0 & 0 \\
        $V^2$ & \textbf{1} & 0 & 0 \\
        $V^3$ & 0 & 0 & 0 \\
    \end{tabular}}
};
\node[above=0.08cm of tab1, font=\scriptsize\bfseries] {Lag $\ell=1$};

\node[below=0.35cm of tab1, text width=5.5cm, font=\footnotesize, align=center] {
    \textbf{(b) Binary lagged adjacency tensor slices} \\
    Entry $\mathbb{A}^{(\ell-1)}_{j,i}=1$ represents $V^i_{t-\ell} \to V^j_t$
};

\end{tikzpicture}
\end{minipage}
\caption{Temporal causal dependencies represented as a (a) lagged causal graph and (b) binary adjacency tensor. Each slice \(\mathbb{A}^{(\ell-1)}\) encodes edges at a discrete lag \(\ell \leq \ell_{\max}\), where entry \(\mathbb{A}^{(\ell-1)}_{j,i}=1\) denotes \(V^i_{t-\ell} \to V^j_t\).}
\label{fig:causal-tensor-combined}
\end{figure}

\paragraph{Adjacency Tensor Representation.} For \(V\) variables and maximum lag \(\ell_{\max}\), the lagged causal graph is equivalently represented by a binary adjacency tensor \(\mathbb{A} \in \{0,1\}^{V \times V \times \ell_{\max}}\), where \(\mathbb{A}^{(\ell)}_{j,i} = 1\) indicates a directed edge \(V^i_{t-\ell} \rightarrow V^j_t\) \cite{stein2024embracing, gkorgkolis2026adversarial}(Figure~\ref{fig:causal-tensor-combined}).

\section{Related Work} \label{sec:related-work}

Despite advances in deep learning, few works explore pre-trained models for Causality. Most approaches remain task or dataset specific and do not exhibit the broad generalization associated with foundation models.

In \textit{causal inference}, \cite{zhang2023towards} propose CInA, a transformer-based framework for estimating average treatment effects (ATEs) \cite{peters2017elements} from observational data, which targets treatment effect estimation rather than causal discovery.

Neural approaches to CD include attention-based models for static graph prediction under i.i.d.\ assumptions \cite{wu2024sample}, supervised sequence formulations in atemporal setups \cite{ke2023learning} and amortized variational methods under Granger causality assumptions \cite{lowe2022amortized}. These methods typically operate on limited synthetic datasets, output summary graphs and do not demonstrate large-scale zero-shot generalization, particularly in temporal or heterogeneous settings.

The closest work is \cite{stein2024embracing}, which formulates CD as a multi-class classification problem, and introduces various architectural backbones, notably transformers \cite{vaswani2017attention, zhou2021informer}. Although conceptually aligned with our notion for LCMs, it remains a proof of concept; limited to small systems, training relies on narrow synthetic distributions, and \textbf{performance degrading sharply with increasing dimensionality \cite[Figure~3]{stein2024embracing}}.

\paragraph{Positioning of Our Work.}
We study Large Causal Models as robust, scalable FMs for \textit{temporal} CD. Unlike prior work, we emphasize data diversity as a prerequisite for scaling, train on a large heterogeneous corpus, and evaluate zero-shot and out-of-distribution generalization across diverse benchmarks. This positions LCMs as practical alternatives to existing CD methods.

\section{Problem Formulation} \label{sec:problem-formulation}

\begin{figure}[t!]
    \centering
    \includegraphics[width=1\textwidth]{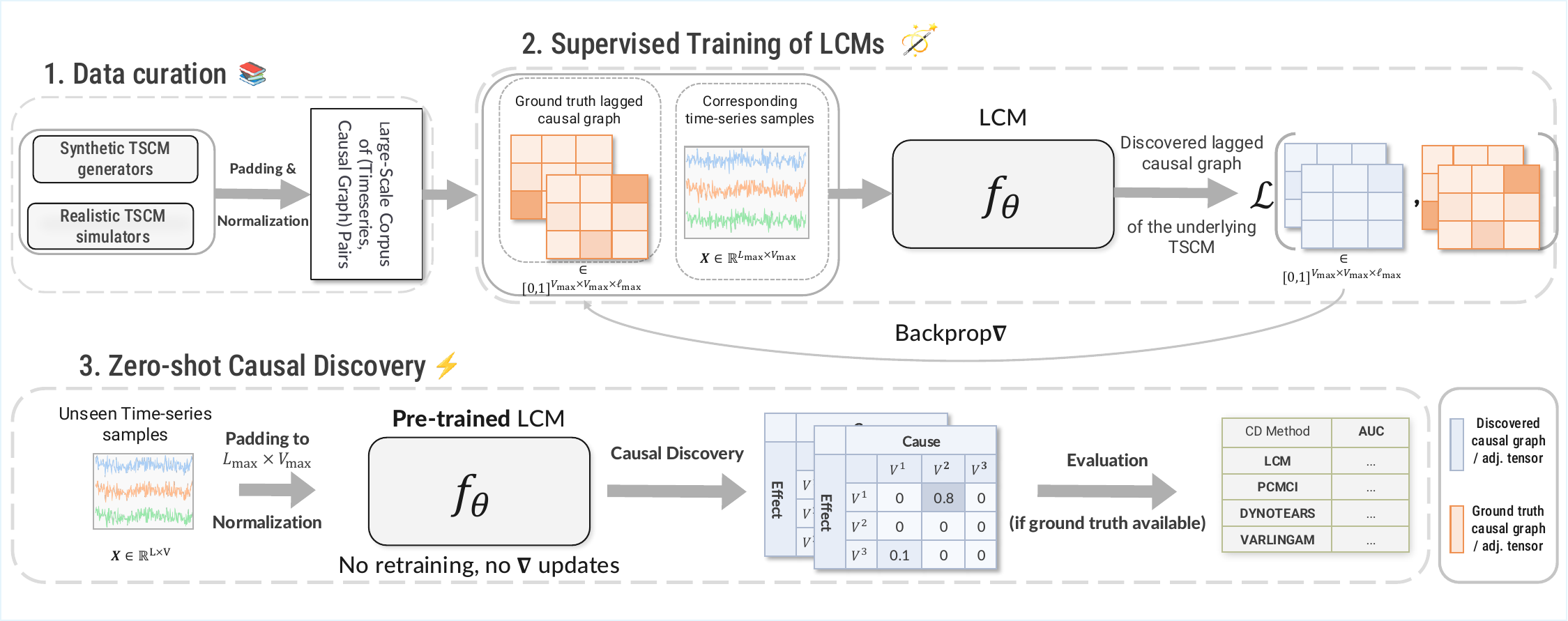}
    \caption{Overview of the large causal model (LCM) pipeline. (1) Synthetic and realistic TSCM generators produce training pairs of multivariate time series and their lagged causal graphs. (2) The LCM is trained via supervised learning on these pairs to discover a lagged adjacency tensor \(\hat{\mathbb{A}}\) for a time series \(\mathbf{X} \in \mathbb{R}^{L \times V}\), padded and normalized for stability. (3) At inference (CD phase), the pre-trained LCM predicts causal structure on unseen datasets in a zero-shot manner.}
    \label{fig:combined}
\end{figure}

We study \textit{temporal CD across heterogeneous datasets and underlying causal mechanisms}. The task is to construct a \textit{Large Causal Model (LCM)}, i.e. an FM that infers lagged causal graphs from multivariate time series, while generalizing across domains, lengths, and causal mechanisms.

Let \( \mathbf{X} = \{\mathbf{X}_t\}_{t=1}^{L} \in \mathbb{R}^{L \times V} \) denote a multivariate time series with \(V\) variables and \(L\) timesteps. Each dataset \(\mathcal{D} = \mathbf{X}\) is generated by a TSCM \(\mathcal{G}\), \(\mathcal{D} \sim \mathcal{P}_{\mathcal{G}}\), where \(\mathbb{A}\) is the (ground truth) lagged adjacency tensor of \(\mathcal{G}\). The task is to infer \(\mathcal{G}\) from \(\mathcal{D}\) via a parametric mapping \(\mathbf{X} \xrightarrow{f_\theta} \hat{\mathbb{A}} \), 
where \(f_\theta\) is a neural model and \(\hat{\mathbb{A}} \in \mathbb{R}^{V_{\max} \times V_{\max} \times \ell_{\max}}\) encodes the discovered lagged causal graph (Figure~\ref{fig:causal-tensor-combined}). Each entry \(\hat{\mathbb{A}}_{j,i,\ell_{\max}-\ell}\) represents the confidence in \(X^i_{t-\ell} \rightarrow X^j_t\).

Probabilistically, the LCM learns an amortized approximation of the conditional distributions \(\mathbb{P}(\mathbb{A}_{j,i,\ell_{\max}-\ell}=1|\mathbf{X})\) under the training distribution over the TSCMs. Rather than performing dataset-specific DAG search, the model learns a parametric approximation of the mapping from time-series samples to causal graphs under the training distribution.
The discovered lagged adjacency tensor hence represents edge probabilities under the implicit prior induced by the training corpus. An overview of the pipeline is depicted in Figure \ref{fig:combined}. 

\paragraph{Model Hyperparameters.} \(L_{\max}\) (max time-series length) and \(V_{\max}\) (maximum input variables) are model-specific, akin to token limits in language models \cite{brown2020language}, while \(\ell_{\max}\) (max assumed lag) follows standard assumptions in temporal CD \cite{runge2018causal}.

\section{Model Overview} \label{sec:model}

\begin{figure}[t!]
  \centering
  \includegraphics[width=0.8\textwidth]{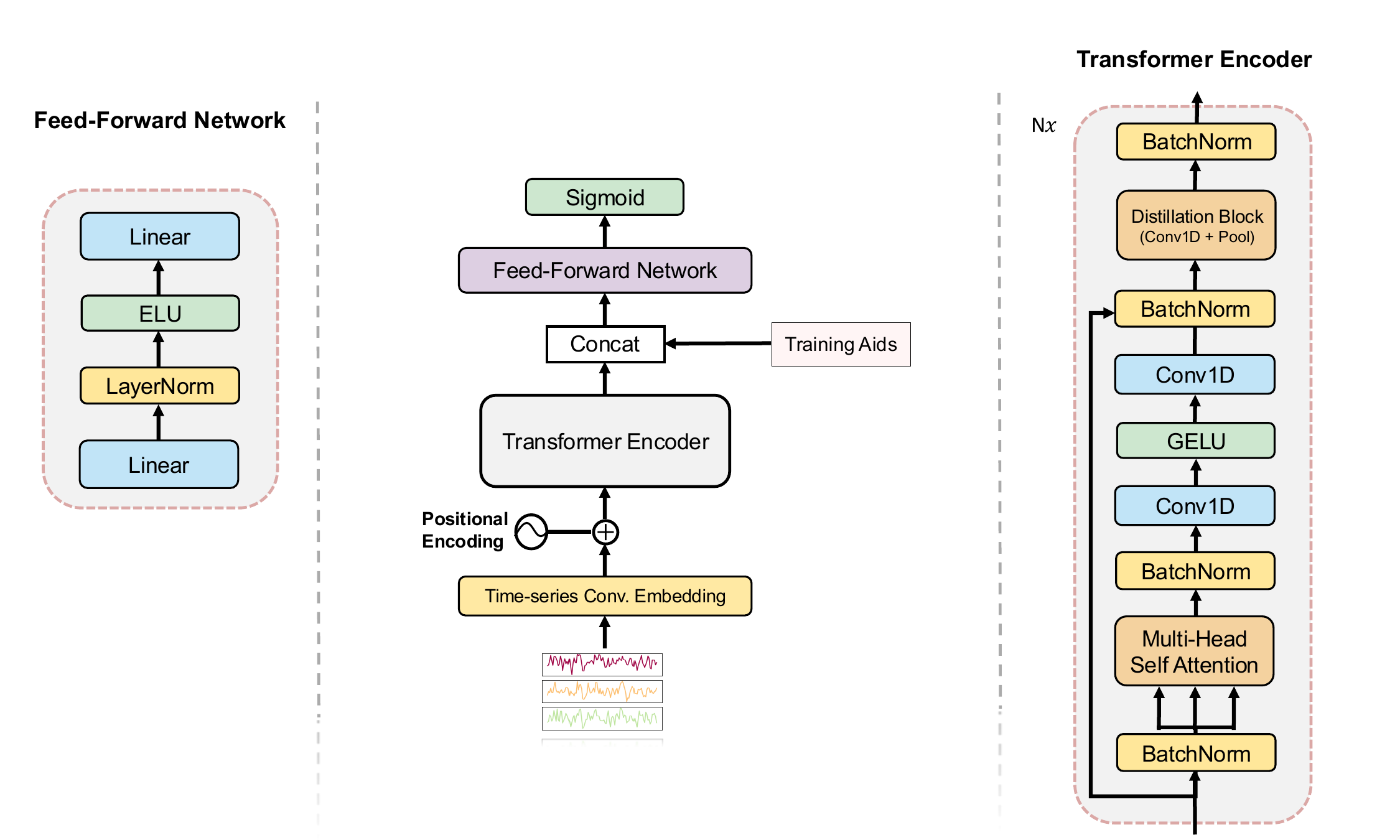}
  \caption{LCM architecture: A multivariate time series is embedded via Conv1D layers and positional encodings, processed through a Transformer encoder stack with distillation blocks, and augmented with lagged cross-correlations (training aids). A feedforward head outputs a lagged adjacency tensor representing the discovered temporal causal graph. Adapted from \cite{kougioulis2025large}.}
  \label{fig:main-arch}
\end{figure}

The implemented LCM, illustrated in Figure~\ref{fig:main-arch}, follows a 
convolution-enhanced Transformer encoder \cite{stein2024embracing, zhou2021informer}:

\paragraph{Input Embeddings.} Input samples \(\mathbf{X} \in \mathbb{R}^{L \times V}\) 
are normalized and noise-padded (Appendix~\ref{app:input-padding}) to 
\(\mathbb{R}^{L_{\max} \times V_{\max}}\), projected via Conv1D to extract local 
temporal features at dimension \(d_\text{model}\), with positional 
encodings added element-wise \cite{vaswani2017attention}.

\paragraph{Encoder Stack.} \(N\) layers of multi-head self-attention and feedforward 
networks with residual connections produce contextualized representations. A 
Conv1D-based distillation layer aims to reduces sequence length for long-horizon processing.

\paragraph{Training Aids.} Lagged cross-correlations are concatenated to the final 
encoder representations to enhance inductive bias (\textit{correlation injection} 
\cite{stein2024embracing}).

\paragraph{Feedforward Head.} A fully-connected network, followed by a sigmoid activation outputs 
\(\hat{\mathbb{A}} \in [0,1]^{V_{\max} \times V_{\max} \times \ell_{\max}}\), where 
\(\mathbb{A}_{j,i,\ell}\) is the probability of \(X^i_{t-\ell} \rightarrow X^j_t\).

The framework is encoder-agnostic; we adopt Transformers as a scalable, well-understood backbone for multivariate time series \cite{liu2024atimer, liu2024btimer, das2024decoder} rather than as a novel contribution; details are provided in Appendix~\ref{app:lcm-architecture}.

\subsection{Loss Function} \label{subsec:loss}

Supervised training of LCMs requires an informative objective that encourages accurate edge predictions while leveraging any observed statistical dependencies. A composite loss is employed, combining supervised edge prediction with a correlation-based regularizing term.

\paragraph{Edge Prediction Loss.} 
Let \(\hat{\mathbb{A}} \in \mathbb{R}^{V_{\max} \times V_{\max} \times \ell_{\max}}\) denote the discovered lagged adjacency tensor and \(\mathbb{A}\) the ground truth. Each potential edge \(X^i_{t-\ell} \rightarrow X^j_t\) is treated as a binary classification problem, optimized via binary cross-entropy:
\[
\mathcal{L}_{\text{edge}} 
= \frac{1}{V_{\max}^2 \ell_{\max}} 
\sum_{i,j,\ell} \mathrm{BCE}\!\left( \mathbb{A}_{j,i,\ell}, \hat{\mathbb{A}}_{j,i,\ell} \right)
\]

\paragraph{Regularization term.} 
As observed by \cite{stein2024embracing}, causal discovery FMs for time-series can benefit from the inclusion of observed statistics. To improve stability and generalization, an auxiliary term is added, aligning predictions with lagged cross-correlations observed in input time-series. For \(\mathbf{X} = \{\mathbf{X}_t\}_{t=1}^L\), correlations up to lag \(\ell_{\max}\) are computed and normalized (we denote this tensor \(\tilde{\mathrm{CC}}\)). The correlation loss corresponds to a weighted mean squared error: \(\mathcal{L}_{\text{corr}} = \mathbb{E} \big[ (\hat{\mathbb{A}} - \tilde{\mathrm{CC}})^2 \odot \tilde{\mathrm{CC}}^\gamma \big]\) where \(\gamma > 1\) emphasizes strong correlations. This term encourages confident causal predictions to align with empirical temporal dependencies without imposing hard constraints. Although \cite{stein2024embracing} adopt a different regularizing term, we adopt their terminology, \textit{correlation regularization (CR)}, for clarity.

\paragraph{Overall Objective.} 
The final loss combines the two components: \(
\mathcal{L} = \lambda_{\text{edge}} \, \mathcal{L}_{\text{edge}} + \lambda_{\text{corr}} \, \mathcal{L}_{\text{corr}}
\), where \(\lambda_{\text{edge}} = 1\) and \(\lambda_{\text{corr}} = 3/4\) are selected by default (Table \ref{tab:training_aids_ablation}).

\section{Training Setup} \label{sec:training-setup}

\subsection{Datasets} \label{subsec:datasets}

LCMs require paired samples \((\mathcal{D}, \mathcal{G})\) for supervised training. To promote zero-shot generalization under distribution shift, we construct a large, diverse corpus combining \textit{synthetic}, \textit{semi-synthetic}, and \textit{realistic} time-series datasets.

\subsubsection{Synthetic.} \label{subsubsec:synthetic}
We generate random TSCMs by sampling graphs from parametrized random families (e.g., Erd\H{o}s--R\'enyi \cite{hagberg2007exploring, fienberg2012brief, brouillard2020differentiable}) and performing ancestral sampling \cite[Section~4.2.5]{murphy2023probabilistic}, dynamically varying the number of variables, \(\ell_{\max}\), graph density, and functional mechanisms (linear and nonlinear additive noise models \cite{kalainathan2022structural}). Our primary large-scale collection (\texttt{Synthetic\_2}) contains \(230\)k instances; \texttt{CDML} \cite{lawrence2020cdml} and \texttt{Synthetic\_1} \cite{stein2024embracing} provide additional synthetic evaluation. Details in Appendix~\ref{app:datasets}.

\subsubsection{Semi-synthetic.} \label{subsubsec:semi-synthetic}
For Out-Of-Distribution (OOD) evaluation, we use mechanistic simulators with known causal connectivities: \(f\)MRI-based dynamic nonlinear models (\texttt{\(f\)MRI\_5}, \texttt{\(f\)MRI}) \cite{buxton1998dynamics, smith2011network, lowe2022amortized} and Kuramoto coupled oscillator systems (\texttt{Kuramoto\_5}, \texttt{Kuramoto\_10}) \cite{kuramoto1975self, lowe2022amortized}. These are never used for training and serve exclusively as OOD benchmarks.

\subsubsection{Realistic.} \label{subsubsec:realistic} We incorporate realistic TSCMs derived from real-world time series following \cite{gkorgkolis2026adversarial}, to the best of our knowledge the only method constructing calibrated causal twins from observed, real data. Input datasets span multiple domains (energy, weather, transportation) \cite{hahn2023time}, yielding 45k instances. Additional realistic datasets are further reserved for OOD evaluation (Appendix~\ref{app:datasets}).

\paragraph{Mixture Collections.} To analyze synthetic-to-realistic ratio effects, we construct four mixture datasets (\texttt{MIX\_1}--\texttt{MIX\_4}) with proportions from 100/0 to 20/80, each containing \(50\)k instances with identical temporal characteristics.

\paragraph{Large-Scale Training Corpus.} Our main corpus (\texttt{LS}) combines synthetic and realistic data, totaling 275k instances (\textbf{137.5M time points}). All evaluations use holdout test sets for in-distribution scenarios and separate OOD benchmarks. Beyond LCM training, these corpora \textbf{constitute a substantial benchmark resource} for systematic evaluation of existing and future temporal CD methods.

\section{Experimental Setup} \label{sec:exp-setup}

Dataset configurations are described in Section \ref{subsec:datasets}. Regarding LCMs, we consider multiple variants with increasing model capacity, from \(\approx 900\)K to \(24\)M parameters. Larger models are used to study scalability in variable count and data size, while smaller models are employed for ablation studies and comparisons (Subsections \ref{subsec:exp-correlations}, \ref{subsec:exp-synth-sim}). The selection of such parameter counts is motivated by established time-series forecasting FMs, such as MOIRAI \cite{woo2024unified}, Timer \cite{liu2024atimer}, and TimesFM \cite{das2024decoder}. Configurations, implementation and training details are provided in Appendix \ref{app:model-sizes}.

\subsection{Evaluation Metrics} \label{sec:cd-metrics}

Edge discovery performance is evaluated by the \textit{Area Under the ROC Curve (AUC)} \cite{bradley1997use}, computed directly from the discovered lagged adjacency tensor \cite{cheng2023causaltime, gkorgkolis2026adversarial}: edges are ranked by confidence, and the area under the TPR–FPR curve is computed across thresholds. Higher AUC scores indicate better recovery of the ground-truth lagged causal graph. Standard errors are estimated across datasets within each collection.

Statistical significance of AUC differences between LCMs and baseline methods is asserted using the Wilcoxon signed-rank test \cite{conover1999practical}, a non-parametric paired test appropriate for potentially non-Gaussian AUC distributions, under a Bonferroni correction (\(\alpha_{\text{corrected}} = \alpha/k\)) where \(k\) is the number of performed comparisons and \(\alpha\) the level of significance.

\subsection{Baselines} \label{subsec:baselines}

We compare LCMs against established temporal CD methods: \textbf{PCMCI} \cite{runge2019inferring}, \textbf{DYNOTEARS} \cite{pamfil2020dynotears} and \textbf{VARLinGAM} \cite{hyvarinen2010estimation}, representing constraint-based, score-based, and functional-model approaches respectively. For small datasets (\(V=5\), \(\ell_{\max}=3\)), we additionally include the pre-trained Transformer from \cite{stein2024embracing} (\(\approx 1.4\text{M}\) parameters).

All baselines use official implementations. Outputs are converted to lagged adjacency tensors for fair comparison. Regarding constraint-based methods, edge confidence corresponds to inverse p-values. For methods producing causal effect estimates, edge probabilities are estimated via bootstrap resampling (Appendix \ref{app:baselines} \& Algorithm \ref{alg:bootstrap}).

\section{Experimental Results} \label{sec:results}

We present representative results in each of the following subsections. Our experimental evaluation spans a broad range of settings, including both in-distribution (holdout, out-of-sample test sets) and out-of-distribution scenarios. Across these settings, we show that LCMs consistently match or outperform classical baselines, which require fitting a new model for each individual dataset. This further highlights the effectiveness of LCMs for efficient CD. 

\subsection{Observed Statistics Improve LCM Performance} \label{subsec:exp-correlations}

We study the impact of incorporating observed statistics into LCM training through a set of \textit{training aids}, as defined in Section \ref{sec:model}. We evaluate a baseline LCM, LCM + CI, and LCM + CI + CR following a grid search procedure \cite{hutter2019automated} over \(\lambda_{\text{corr}} \in \{1/4, 1/2, 3/4, 1\}\), keeping all other hyperparameters fixed. Experiments are conducted on the \texttt{Synthetic\_1} holdout set.


\begin{table}[h!]
\centering
\caption{Ablation of training aids on \texttt{Synthetic\_1}. Asterisks (*) indicate statistical significance over the preceding model (left column).}
\label{tab:training_aids_ablation}
\footnotesize
\setlength{\tabcolsep}{3pt}
\begin{tabular}{lcccccc}
\toprule
 & LCM & +CI & \multicolumn{4}{c}{+CR ($\lambda$)} \\
\cmidrule(lr){4-7}
 &  &  & .25 & .5 & .75 & 1.0 \\
\midrule
AUC
& .868
& \(.914^{\textbf{*}}\)
& \(.926^{\textbf{*}}\)
& \(.918^{\textbf{*}}\)
& \textbf{.926}
& \(.925\) \\
\bottomrule
\end{tabular}
\end{table}

Using these training aids substantially improves performance. CI alone yields a significant improvement over the baseline, and adding CR further increases AUC across all tested weights. The highest mean AUC is achieved for \(\lambda_{\text{corr}} \in [0.25, 0.75]\), with no statistically significant difference among them. We adopt \(\lambda_{\text{corr}}=0.75\) for all subsequent experiments. In the remainder of this paper, we refer to the combination of CI and CR collectively as \textit{training aids}.

\subsection{LCMs benefit from training on mixtures of synthetic and realistic data} \label{subsec:exp-synth-sim}

Training LCMs solely on synthetic data limits generalization beyond narrowly defined causal structures. To mitigate this, we adopt a \textit{dataset mixing} approach, combining synthetic with realistic datasets during training (Subsection \ref{subsubsec:realistic}).

Table \ref{tab:res-mixture} shows OOD CD performance on semi-synthetic \(f\)MRI benchmarks and the real-data–derived \texttt{Power} and \texttt{Climate} collections. Models trained solely on synthetic cases are consistently outperformed by mixed-dataset models.  

Adding a modest fraction of realistic data (\(20\%\)) yields systematic, statistically significant performance gains. Higher proportions provide marginal or inconsistent improvements. Thus, an \(80/20\) synthetic-to-realistic mixture balances structural diversity and real-world fidelity, consistent with recent time-series FM findings \cite{das2024decoder}, and is adopted for all subsequent large-scale experiments.

\begin{table}[t!]
\centering
\caption{Out-of-distribution performance of LCMs trained on varying synthetic/realistic mixtures, evaluated on semi-synthetic \(f\)MRI benchmarks and the OOD \texttt{Power} \& \texttt{Climate} benchmarks. Statistical significance versus the \(80/20\%\) reference (\textbf{bold}) is indicated by an asterisk (\textbf{*}).} 
\label{tab:res-mixture}
\small
\renewcommand{\arraystretch}{1.15}
\begin{tabular}{lcccc}
\toprule
\textbf{Synth/Realistic\%} & \textbf{\(f\)MRI\_5} & \textbf{\(f\)MRI} & \textbf{Power} & \textbf{Climate} \\
\midrule
100/0\%  & \(.924^{\textbf{*}}\) {\tiny $\pm$ .050} & \(.874^{\textbf{*}}\) {\tiny $\pm$ .142} & \(.966^{\textbf{*}}\) {\tiny $\pm$ .011} & \(.938^{\textbf{*}}\) {\tiny $\pm$ .035} \\
\textbf{80/20\%}  & \textbf{.966} {\tiny $\pm$ .028} & \textbf{.960} {\tiny $\pm$ .028} & \textbf{.981} {\tiny $\pm$ .006} & \textbf{.982} {\tiny $\pm$ .011} \\
50/50\%  & \(.951\) {\tiny $\pm$ .050} & \(.944^{\textbf{*}}\) {\tiny $\pm$ .047} & \(.979\) {\tiny $\pm$ .008} & \(.988\) {\tiny $\pm$ .010} \\
20/80\%  & \(.968\) {\tiny $\pm$ .004} & \(.966^{\textbf{*}}\) {\tiny $\pm$ .036} & \(.975^{\textbf{*}}\) {\tiny $\pm$ .010} & \(.986\) {\tiny $\pm$ .015} \\
\bottomrule
\end{tabular}
\end{table}

\subsection{Rich Training Data Enable Scalable, Large-Scale LCMs} \label{subsec:res-lcms}

Prior work suggests that \textit{scaling LCMs to higher-dimensional settings leads to near-random performance, even with increased model capacity \cite[Figure~3]{stein2024embracing}, implying the need for prohibitively deep models}. Our results indicate that these scaling failures are inherent to limitations of the training data distribution rather than architectural constraints. By training on rich datasets, LCMs can scale beyond previously studied small systems (e.g., 3–5 variables), demonstrating stable convergence and generalization up to \(V_{\max}=12\) in our experiments.

\subsubsection{Performance on Synthetic Data} \label{subsec:res-in-dist}

Table \ref{tab:res-in-dist} shows that LCMs achieve state-of-the-art performance on synthetic benchmarks, outperforming all baselines. We attribute their lower performance to their restrictive assumptions, as our models benefit from exposure to a diverse large-scale training corpus.

\begin{table}[h!]
\centering
\caption{Causal discovery performance (AUC) of large-scale LCMs and baseline methods across \textit{synthetic} benchmarks. Results are reported as mean \(\pm\) standard deviation across datasets. An asterisk indicates a statistically significant difference with respect to the best-performing model (LCM-\texttt{12.2M}), which achieves the highest mean AUC across all semi-synthetic collections.}
\label{tab:res-in-dist}
\small
\renewcommand{\arraystretch}{1.15}
\begin{tabular}{lccc}
\toprule
\textbf{Model} & \textbf{Synthetic\_1} & \textbf{Synthetic\_2} & \textbf{CDML} \\
\midrule
\(\text{LCM}_{\texttt{2.5M}}\) & .995\(\textbf{*}\) {\tiny $\pm$ .013} & .903\(\textbf{*}\) {\tiny $\pm$ .107} & .772 {\tiny $\pm$ .177}\\
\(\text{LCM}_{\texttt{9.4M}}\) & .995\(\textbf{*}\) {\tiny $\pm$ .013} & .906\(\textbf{*}\) {\tiny $\pm$ .106} & .768 {\tiny $\pm$ .171} \\
\(\text{LCM}_{\texttt{12.2M}}\) & .\textbf{996} {\tiny $\pm$ .013} & \textbf{.909} {\tiny $\pm$ .103} & \textbf{.773} {\tiny $\pm$ .172} \\
\(\text{LCM}_{\texttt{24M}}\) &  .990\(\textbf{*}\) {\tiny $\pm$ .012} & .865\(\textbf{*}\) {\tiny $\pm$ .133} & .727\(\textbf{*}\) {\tiny $\pm$ .183} \\
CP (Stein et al.) & .943\(\textbf{*}\) {\tiny $\pm$ .151} & -- & -- \\
PCMCI & .671\(\textbf{*}\) {\tiny $\pm$ .444} & .800\(\textbf{*}\) {\tiny $\pm$ .181} & .521\(\textbf{*}\) {\tiny $\pm$ .108} \\
DYNO & .551\(\textbf{*}\) {\tiny $\pm$ .216} & .605\(\textbf{*}\) {\tiny $\pm$ .152} & 490\(\textbf{*}\) {\tiny $\pm$ .087} \\
VARLiNGAM & .922\(\textbf{*}\) {\tiny $\pm$ .145} & .861\(\textbf{*}\) {\tiny $\pm$ .136} & .517\(\textbf{*}\) {\tiny $\pm$ .099} \\
\bottomrule
\end{tabular}
\end{table}

\subsubsection{Out-of-distribution / Zero-shot Performance} \label{subsec:res-ood}

We evaluate zero-shot transfer on semi-synthetic and real-world datasets (Subsection \ref{subsec:datasets}), unseen during training, to assess generalization under distribution shifts, in Tables \ref{tab:performance-zero-shot-semisynth} \& \ref{tab:performance-zero-shot-realistic} respectively. An asterisk indicates a statistically significant difference with respect to the best-performing model (\(\text{LCM}_{\texttt{24M}}\) for Table \ref{tab:performance-zero-shot-semisynth}, \(\text{LCM}_{\texttt{2.5M}}\) for Table \ref{tab:performance-zero-shot-realistic}), achieving the highest mean AUC across most collections.

\begin{table*}[h!]
\centering
\caption{Out-of-distribution (zero-shot) causal discovery performance (AUC) of large-scale LCMs and baseline methods across \textit{semi-synthetic} benchmarks. Results are reported as mean \(\pm\) standard deviation across datasets.}
\label{tab:performance-zero-shot-semisynth}
\small
\renewcommand{\arraystretch}{1.15}
\begin{tabular}{lcccc}
\toprule
\textbf{Model} & \textbf{\(f\)MRI\_5} & \textbf{\(f\)MRI} & \textbf{Kuramoto\_5} & \textbf{Kuramoto\_10} \\
\midrule
\(\text{LCM}_{\texttt{2.5M}}\) & .981 {\tiny $\pm$ .017} & .978 {\tiny $\pm$ .017} & .956\(\textbf{*}\) {\tiny $\pm$ .017} & .919\(\textbf{*}\) {\tiny $\pm$ .018} \\
\(\text{LCM}_{\texttt{9.4M}}\) & .983 {\tiny $\pm$ .007} & .979 {\tiny $\pm$ .012} & .972 {\tiny $\pm$ .009} & .924\(\textbf{*}\) {\tiny $\pm$ .008} \\
\(\text{LCM}_{\texttt{12.2M}}\) & .986 {\tiny $\pm$ .007} & .982 {\tiny $\pm$ .012} & .953\(\textbf{*}\) {\tiny $\pm$ .016} &.921\(\textbf{*}\) {\tiny $\pm$ .014} \\
\(\text{LCM}_{\texttt{24M}}\) & \textbf{.987} {\tiny $\pm$ .006} & \textbf{.984} {\tiny $\pm$ .010} & \textbf{.973} {\tiny $\pm$ .007} & \textbf{.931} {\tiny $\pm$ .010} \\
CP (Stein et al.) & .883\(\textbf{*}\) {\tiny $\pm$ .078} & -- & .541\(\textbf{*}\) {\tiny $\pm$ .108} & -- \\
PCMCI & .746\(\textbf{*}\) {\tiny $\pm$ .111} & .768\(\textbf{*}\) {\tiny $\pm$ .110} & .476\(\textbf{*}\) {\tiny $\pm$ .103} & .635\(\textbf{*}\) {\tiny $\pm$ .046} \\
DYNO & .528\(\textbf{*}\) {\tiny $\pm$ .152} & .483\(\textbf{*}\) {\tiny $\pm$ .159} & .498\(\textbf{*}\) {\tiny $\pm$ .077} & .519\(\textbf{*}\) {\tiny $\pm$ .027} \\
VARLiNGAM & .793\(\textbf{*}\) {\tiny $\pm$ .095} & .781\(\textbf{*}\) {\tiny $\pm$ .101} & .479\(\textbf{*}\) {\tiny $\pm$ .105} & .647\(\textbf{*}\) {\tiny $\pm$ .053}\\
\bottomrule
\end{tabular}
\end{table*}

\begin{table*}[h!]
\centering
\caption{Out-of-distribution (zero-shot) causal discovery performance (AUC) of 
large-scale LCMs and baseline methods across \textit{realistic} benchmarks. 
Results are mean $\pm$ std.\ across datasets. AQ = AirQualityMS, Clim = Climate, 
Gar = Garments, ETT = ETTm2, Gear = Gearbox, Pow = Power.}
\label{tab:performance-zero-shot-realistic}
\small
\renewcommand{\arraystretch}{1.05}
\setlength{\tabcolsep}{5pt}
\begin{tabular}{lcccccc}
\toprule
\textbf{Model} 
    & \textbf{AQ} 
    & \textbf{Clim} 
    & \textbf{Gar} 
    & \textbf{ETT} 
    & \textbf{Gear} 
    & \textbf{Pow} \\
\midrule
\(\text{LCM}_{\texttt{2.5M}}\)
    & \(\mathbf{.985}_{\pm.014}\) & \(\mathbf{.989}_{\pm.007}\) & \(.995_{\pm.004}\)
    & \(\mathbf{.977}_{\pm.007}\) & \(.998_{\pm.007}\) & \(.980^*_{\pm.007}\) \\
\(\text{LCM}_{\texttt{9.4M}}\)
    & \(.982_{\pm.013}\) & \(.983_{\pm.012}\) & \(.996_{\pm.004}\)
    & \(.965_{\pm.024}\) & \(.998_{\pm.005}\) & \(\mathbf{.983}_{\pm.005}\) \\
\(\text{LCM}_{\texttt{12.2M}}\)
    & \(.977^*_{\pm.018}\) & \(.987_{\pm.010}\) & \(\mathbf{.996}_{\pm.003}\)
    & \(.975_{\pm.010}\) & \(\mathbf{.999}_{\pm.003}\) & \(.983^*_{\pm.005}\) \\
\(\text{LCM}_{\texttt{24M}}\)
    & \(.970^*_{\pm.020}\) & \(.982_{\pm.014}\) & \(.995_{\pm.006}\)
    & \(.972_{\pm.053}\) & \(.996_{\pm.010}\) & \(.979^*_{\pm.008}\) \\
\midrule
PCMCI
    & \(.565^*_{\pm.209}\) & \(.927_{\pm.096}\) & \(.972^*_{\pm.053}\)
    & \(.898^*_{\pm.035}\) & \(.980_{\pm.040}\) & \(.966_{\pm.028}\) \\
DYNO
    & \(.706^*_{\pm.218}\) & \(.708^*_{\pm.118}\) & \(.536^*_{\pm.051}\)
    & \(.566^*_{\pm.037}\) & \(.693^*_{\pm.083}\) & \(.551^*_{\pm.053}\) \\
VARLiNGAM
    & \(.539^*_{\pm.146}\) & \(.938^*_{\pm.075}\) & \(.984^*_{\pm.020}\)
    & \(.906^*_{\pm.024}\) & \(.995^*_{\pm.009}\) & \(.941^*_{\pm.030}\) \\
\bottomrule
\end{tabular}
\end{table*}

Results demonstrate that all model sizes achieve superior or competitive zero-shot performance compared to baseline methods. This highlights that \textit{richer, diverse training data enable strong generalization}, confirming that scaling failures stem from insufficient training distributions rather than architectural limitations.

\subsection{Running Times}

\begin{figure}[t!]
\centering
\includegraphics[width=0.65\textwidth]{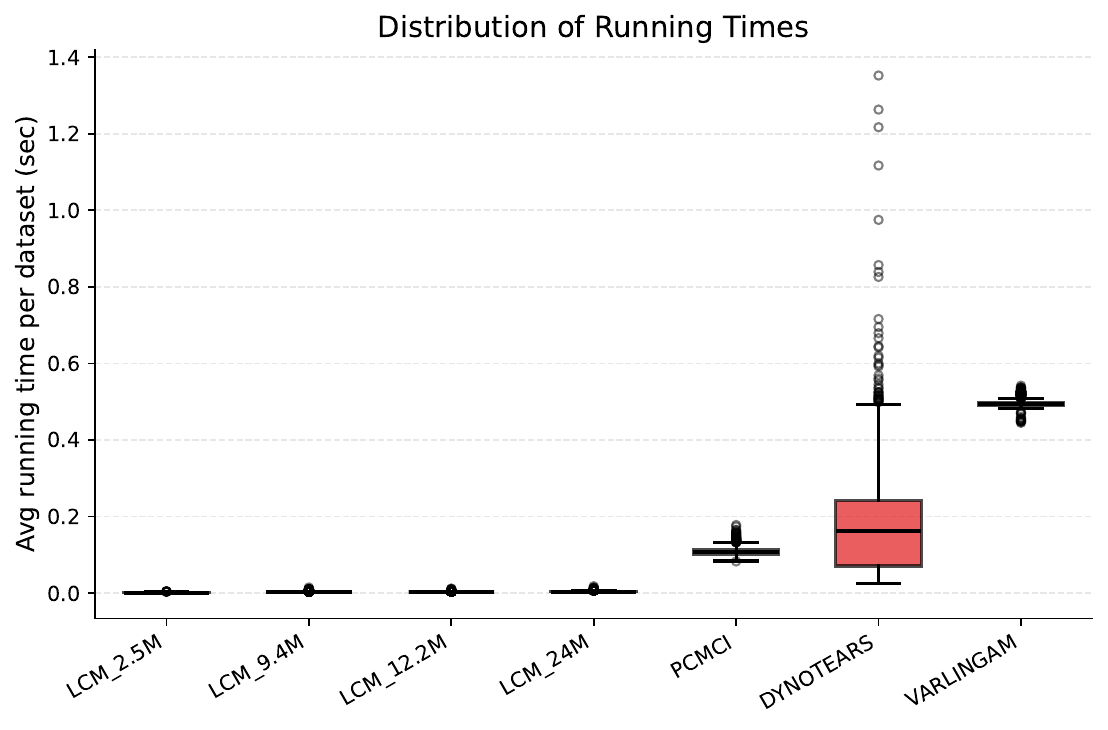}
\caption{Running times (in seconds) for LCMs and baseline algorithms on the \texttt{Synthetic\_2} holdout set, averaged over 10 runs. Traditional methods (e.g., PCMCI \& DYNOTEARS) scale superlinearly with lag and variable count, while Transformer-based LCMs remain effectively independent of input dimensionality due to their constant-time forward pass.} 
\label{fig:runtimes}
\end{figure}

We compare computational runtimes of LCMs and conventional CD algorithms (Section \ref{subsec:baselines}) to highlight scalability advantages. For an LCM, this corresponds to a single forward pass, effectively replacing iterative search of classic CD methods with parametric approximation. Each method is run 10 times per dataset, and results are averaged (Figure \ref{fig:runtimes}).  

A key observation is that LCM runtimes remain effectively \textit{independent of input dimensionality}, as computations operate on learned representations rather than fitting a new model per input dataset. In contrast, traditional methods exhibit superlinear scaling with variable count and temporal lag, possibly limiting applicability in real-time, high-dimensional settings.

\section{Model Complexity} \label{sec:model-complexity}

We derive a closed-form expression for the number of trainable parameters of our LCM models. Let \(B\) denote the number of encoder blocks, \(d_{\mathrm{model}}\) the model dimension, \(n_{\mathrm{heads}}\) the number of attention heads, and \(d_{\mathrm{ff}}\) the feedforward dimension. We introduce the binary indicators \(I_{\mathrm{train}}, I_{\mathrm{distil}} \in \{0,1\}\) for the use of our correlation-based training aids and attention distillation respectively. The kernel size of the convolutional embedding is denoted by \(k\).

The total number of trainable parameters decomposes into terms for (i) input token embeddings, (ii) \(B\) encoder blocks, (iii) \((B-1)\) distillation layers when enabled, and (iv) the final causal prediction head. Summing the above,

\begin{equation*}
\label{eq:param-count}
\begin{aligned}
P_{\mathrm{total}} &= (d_{\mathrm{model}} V_\text{max} \cdot k + d_{\mathrm{model}}) \\
&\quad + B\bigl(4 \cdot d_{\mathrm{model}}^2 + 2 d_{\mathrm{ff}} \cdot d_{\mathrm{model}} + d_{\mathrm{ff}} + 9 \cdot d_{\mathrm{model}}\bigr) \\
&\quad + I_{\mathrm{distil}}(B-1)\bigl(d_{\mathrm{model}}^2 \cdot k + 3 \cdot d_{\mathrm{model}}\bigr) \\
&\quad + \bigl((d_{\mathrm{model}} + I_{\mathrm{train}} \cdot A)d_{\mathrm{ff}} + d_{\mathrm{ff}}\bigr) + (d_{\mathrm{ff}} \cdot A + A).
\end{aligned}
\end{equation*}

where \(A = V_\text{max}^2 \ell_\text{max}\) and \(d_{\mathrm{model}}\) is assumed to be divisible by \(n_{\mathrm{heads}}\). Dominant term grows as \(\Theta(V_\text{max}^2 \ell_\text{max})\), i.e., pairwise causal edges in the feedforward head. For \(V_\text{max}=25, \ell_\text{max}=3\) with \texttt{24M} hyperparameters, the head alone contributes \(\approx 2\text{M}\) parameters, total \(\approx 75\text{M}\). Our considered input dimensionality (\(V_\text{max}=12, \ell_\text{max}=3\)) represents a careful balance between scalability, generalization, and applicability, with higher-dimensional extensions possible via time-series selection \cite{vareille2024chronoepilogi}, for example in genomics applications.

\section{Training Data–Model Size Convergence}

We study the interaction between model capacity and training data size by training 
500K, 1M, and 2M parameter LCMs on subsampled datasets of sizes \(\{10\text{K}, 25\text{K}, 50\text{K}, 100\text{K}\}\) from our large-scale corpus (five random seeds; fixed validation/test sets of \(2\cdot10^3\) samples).

\begin{figure}[t!]
\centering
\includegraphics[width=0.5\textwidth]{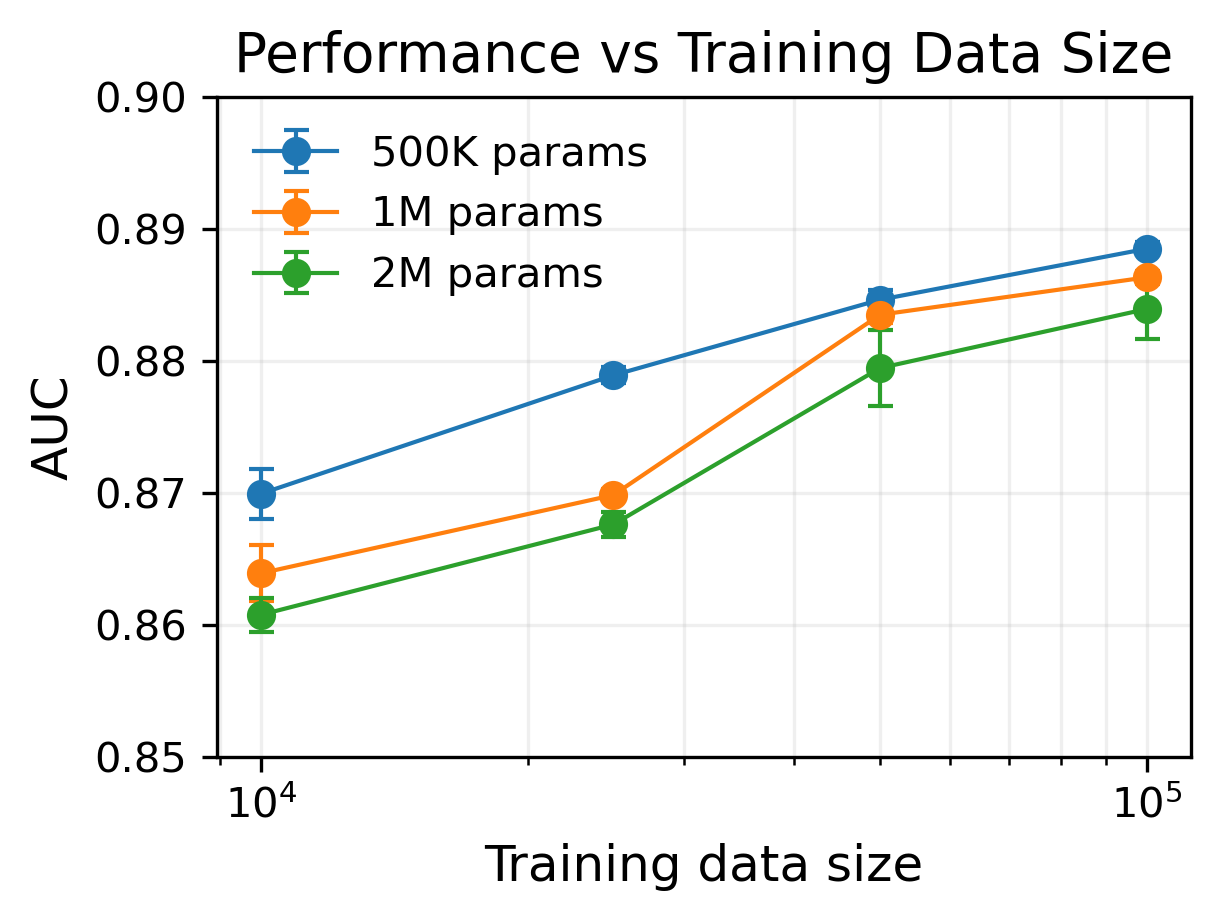}
\caption{Empirical convergence of LCMs with increasing training data. Test AUC for 
500K, 1M, and 2M parameter models trained on subsampled datasets. Validation/test 
sets are fixed to isolate the effect of data scale.}
\label{fig:td-conv}
\end{figure}

As shown in Figure~\ref{fig:td-conv}, smaller models saturate early with diminishing returns from additional data, while larger models continue to improve at scale. Results confirm that LCM performance is jointly constrained by model capacity and data availability: scaling either alone yields limited gains, but scaling both together leads to consistent improvements. We do not claim a universal scaling law; rather, these findings inform practical design choices for large-scale temporal CD.

\section{Concluding Remarks \& Future Work}

\textit{Large Causal Models} (LCMs) are foundation models for temporal causal discovery that amortize CD under a supervised scheme, enabling fast, single-pass graph prediction without any optimization. Trained on mixed data collections, LCMs exhibit strong in- and out-of-distribution generalization, consistently matching or outperforming established baselines. Future work includes relaxing causal sufficiency to handle latent confounders, contemporaneous 
effects, and scaling to higher-dimensional systems and longer horizons.

\section{Limitations}
LCMs operate under the causal assumptions defined in Section~\ref{sec:introduction} and do not detect or correct violations. When assumptions are violated, predictions may reflect training distribution biases rather than true causal structure, and should be interpreted accordingly.

\section{Impact Statement}
LCMs enable scalable, zero-shot temporal causal discovery without dataset-specific retraining, potentially accelerating scientific discovery in fields such as biology and marketing by reducing reliance on costly randomized control trials \& A/B testing.

\subsubsection{\discintname}
Nikolaos Kougioulis, Nikolaos Gkorgkolis \& Ioannis Tsamardinos have received funding from Huawei Ireland Research Centre, Dublin, Ireland.

\bibliographystyle{splncs04}
\bibliography{references}

\newpage

\appendix

\subsection{Causal Assumptions \& Brief Definitions} \label{app:definitions}

All Causal Discovery algorithms are governed by a set of assumptions, regarding the statistical properties of the data and the underlined structure of the causal model. The following assumptions are defined in a way to be applied to either a (standard) causal graph or a temporal causal graph.

\begin{definition}[Causal Markov Condition, \cite{spirtes2001causation, pearl2009causality}] Let \(\mathcal{G}\) be a causal graph with vertex set \(\mathbf{V}\) and \(\mathbb{P}\) a probability distribution over \(\mathbf{V}\), generated by the causal structure induced by \(\mathcal{G}\). Then \(\mathcal{G}\) and \(\mathbb{P}\) satisfy the Markov Condition if and only if \(~\forall ~W \in \mathbf{V}\), \(W\) is independent of its non-descendants on the causal DAG (non causal effects) given its parents (direct causes) on \(\mathcal{G}\).
\end{definition}

\begin{definition}[Faithfulness, \cite{spirtes2001causation, pearl2009causality}] Let \(\mathcal{G}\) be a causal graph and \(\mathbb{P}\) a probability distribution over \(\mathbf{V}\). We say that \(\mathcal{G}\) and \(\mathbb{P}\) are faithful to each other if and only if the all and only the independence relations of \(\mathbb{P}\) are entailed by the Causal Markov condition of \(\mathcal{G}\). Specifically, \(\mathcal{G}\) and \(\mathbb{P}\) satisfy the Faithfulness Condition if-f every conditional independence relation true in \(\mathbb{P}\) is entailed by the Causal Markov Condition applied to \(\mathcal{G}\).
\end{definition}

\begin{definition}[Causal Sufficiency, \cite{spirtes2001causation, pearl2009causality}] A set \( \mathbf{V} \) of variables is causally sufficient for a population if and only if in the population every common cause of any two or more variables in \( \mathbf{V} \) is in \(V\), or has the same value for all units in the population. The common cause \( Z \) of two or more variables in a DAG \( X \leftarrow Z \rightarrow Y \) is called a confounder of \( X \) and \( Y \). Hence Causal Sufficiency implies no unobserved confounders. The notion of causal sufficiency is being used without explicitly mentioning the population.
\end{definition}

\begin{definition}[Causal Stationarity, \cite{runge2018causal}] Consider the SCM description from Section \ref{sec:introduction}. If the causal relationships between variables \( \left(V^i_{t-\tau}, V^j_t \right) \) for time lag \(\tau>0\) also hold for all time-shifted versions \( \left(V^i_{t' -\tau}, V^j_{t'} \right) \), the described process is \textit{causally stationary}. Informally, the graph structure and noise distribution of the SCM are time-invariant.
\end{definition}

\begin{definition}[Time-series Stationarity, \cite{brockwell1991time}]
Let \( \left\{ X_t \right\}_{t \in \mathbb{Z}} \) be a time-series with index set \( \mathbb{Z} = \left\{ 0, \pm 1, \pm 2, \ldots \right\} \). The series is said to be stationary if:
\begin{enumerate}
    \item \( \mathbb{E}|X_t|^2 < \infty, \quad \forall t \in \mathbb{Z} \)
    \item \( \mathbb{E}X_t = m, \quad \forall t \in \mathbb{Z} \)
    \item \( \gamma_X(r,s) = \gamma_X(r + t, s + t), \quad \forall r, s, t \in \mathbb{Z} \)
\end{enumerate}
where \( \gamma_X(r,s) = \text{Cov}(X_r, X_s) = \mathbb{E} \left[ (X_r - \mathbb{E}X_r)(X_s - \mathbb{E}X_s) \right], ~r, s \in \mathbb{Z} \) is the autocovariance function of the stochastic process \( \left\{ X_t \right\}_{t \in \mathbb{Z}} \).
\end{definition}

We now give definitions on used terminology for our causal graphs. 

\begin{definition}[Lagged Causal Graph - Window Causal Graph]
A \textit{Lagged Causal Graph} is a directed acyclic graph (DAG) that represents the causal relationships between time-shifted variables in a multivariate time-series. Formally, let $\mathbf{V}_t = \{V^1_t, V^2_t, \dots, V^k_t\}$ denote the set of observed variables at time step $t$. A lagged causal graph $\mathcal{G}$ contains directed edges of the form $V^i_{t-\tau} \rightarrow V^j_t$, where $\tau \in \{1, 2, \dots, \ell_{\max}\}$ is a positive time lag. An edge $V^i_{t-\tau} \rightarrow V^j_t$ indicates that $V^i$ is a direct cause of $V^j$ with a time lag of $\tau$.
\end{definition}
 
\begin{definition}[Summary Causal Graph]
 Given a lagged causal graph $\mathcal{G}$ with edges of the form $V^i_{t-\tau} \rightarrow V^j_t$ for various lags $\tau$, the corresponding summary graph $\mathcal{S}$ contains an edge $V^i \rightarrow V^j$ if there exists at least one lag $\tau$ such that $V^i_{t-\tau}$ is a direct cause of $V^j_t$ in $\mathcal{G}$.
\end{definition}

Essentially, a summary graph is a simplified representation of the lagged causal structure in a time-series, where the temporal information about lags is abstracted away. The summary graph captures whether a causal relationship exists between two variables, but it does not specify the lag or delay at which the causal influence occurs. 

\subsection{Input Handling and Padding Strategies} \label{app:input-padding}

\begin{figure}[h]
\centering
\renewcommand{\arraystretch}{1.4}
\setlength{\tabcolsep}{4pt}
\begin{tabular}{
| >{\raggedright\arraybackslash}m{2.6cm}
| >{\centering\arraybackslash}m{3.8cm}
| >{\centering\arraybackslash}m{4.2cm} |
}
\hline
& \textbf{W/o Training Aids}
& \textbf{W/ Training Aids} \\
\hline
\textbf{Input dimensions}
& \(d_{\text{model}}\)
& \(d_{\text{model}} + 2 \cdot V_{\max}^2 \times \ell_{\max}\) \\
\hline
\textbf{Hidden state}
&
\begin{tikzpicture}[scale=0.3]
\coloredblockscontig{0}{3}{softorange}
\end{tikzpicture}
&
\begin{tikzpicture}[scale=0.3]
\coloredblockscontig{0}{3}{softorange}
\coloredblockscontig{3}{3}{softblue}
\end{tikzpicture} \\
\hline
\end{tabular}
\caption{Hidden state representations with and without auxiliary training aids.}
\label{app:input-concat}
\end{figure}

In order to handle variable-length time-series, variable numbers of time-series and differing maximum lags across datasets, we adopt a set of standardized padding strategies, both for the input time-series and the ground truth lagged adjacency tensor. This enables consistent batch processing and generalization of trained models across heterogeneous dataset configurations, as the model is able to handle an input of fixed dimensions, as in conventional foundation models. 

\paragraph{Time-Series Padding.}  
When the number of observed time-steps (samples) \(L\) is less than the configured maximum number of samples \( L_\text{max} \), the time-series is \textit{padded with Gaussian noise \( \mathcal{N}(0,0.01)\) along the time-step dimension}. This prevents introducing zero artifacts, preserving the marginal statistics, and makes the model robust to handling varying sequence lengths. When inputs are longer than the configured maximum (either in time-steps \(L\) or the number of time-series \(d_\text{max}\)), the excess is \textit{truncated} to fit the maximum dimensions. This is conceptually similar to how Foundation models like large language models - LLMs truncate input lengths that exceed the maximum allowed tokens \cite{brown2020language}.

\paragraph{Causal Graph Padding.}  

A similar procedure is carried out for padding of the lagged causal graph. When a dataset has fewer variables than \(V_\text{max} \), or the lagged causal graph assumes a lower maximum lag \( l < \ell_\text{max} \), the lagged adjacency tensor is \textit{zero-padded} alongside the unused variable and lag dimensions. Such padding explicitly encodes the absence of nodes or time dependencies outside the observed configuration. Such a step, together with time-series padding described previously, is crucial for both training of the LCMs, as well as during evaluation against the ground truth causal graph, as it allows handling of time-series and causal graph pairs of variable time-step, node and lag dimensions. An illustration of the above is shown in Figure \ref{app:input-padding}.

\paragraph{Input Concatenation.} Implemented LCMs, unless specified otherwise, concatenate lagged crosscorrelations (training aids), which are flattened along the batch dimension to the last hidden state (as illustrated in Figure \ref{fig:main-arch} of the main text). We depict the above in Figure \ref{app:input-concat}.

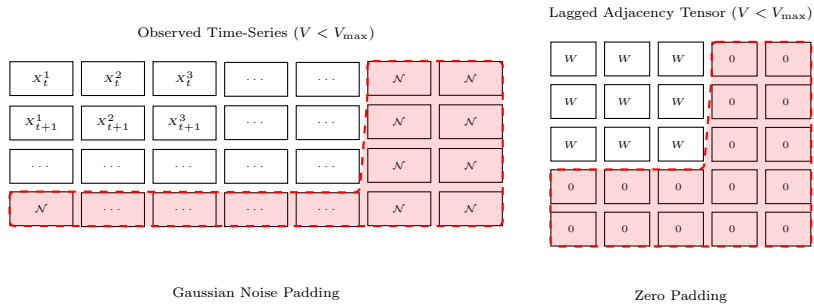
\begin{figure}[h]
\centering
\begin{tikzpicture}[
    every node/.style={scale=0.75},
]
\matrix (A) [matrix of nodes,
    nodes={
        draw, 
        minimum width=10mm, 
        minimum height=6mm,
        align=center,
        text width=9mm,
        font=\tiny
    },
    column sep=1mm, row sep=1mm
] at (0,0)
{
  $X^1_t$ & $X^2_t$ & $X^3_t$ & $\cdots$ & $\cdots$ & $\mathcal{N}$ & $\mathcal{N}$ \\
  $X^1_{t+1}$ & $X^2_{t+1}$ & $X^3_{t+1}$ & $\cdots$ & $\cdots$ & $\mathcal{N}$ & $\mathcal{N}$ \\
  $\cdots$ & $\cdots$ & $\cdots$ & $\cdots$ & $\cdots$ & $\mathcal{N}$ & $\mathcal{N}$ \\
  $\mathcal{N}$ & $\cdots$ & $\cdots$ & $\cdots$ & $\cdots$ & $\mathcal{N}$ & $\mathcal{N}$ \\
};
\node[above=2mm of A-1-4, font=\scriptsize] {Observed Time-Series ($V < V_{\max}$)};
\begin{scope}
\filldraw[
    fill=red,
    fill opacity=0.15,
    draw=red,
    dashed,
    thick,
    rounded corners=2pt
]
(A-1-6.north west) --
(A-1-7.north east) --
(A-4-7.south east) --
(A-4-1.south west) --
(A-4-1.north west) --
(A-4-5.north east) --
(A-1-6.south west) --
cycle;
\end{scope}
\node[below=1.3cm of A-3-4, font=\scriptsize] {Gaussian Noise Padding};

\matrix (B) [matrix of nodes,
    nodes={
        draw,
        minimum width=8mm,
        minimum height=6mm,
        align=center,
        font=\tiny
    },
    column sep=1mm, row sep=1mm,
    right=0.4cm of A
] 
{
  $W$ & $W$ & $W$ & $0$ & $0$ \\
  $W$ & $W$ & $W$ & $0$ & $0$ \\
  $W$ & $W$ & $W$ & $0$ & $0$ \\
  $0$ & $0$ & $0$ & $0$ & $0$ \\
  $0$ & $0$ & $0$ & $0$ & $0$ \\
};
\node[above=2mm of B-1-3, font=\scriptsize] {Lagged Adjacency Tensor ($V < V_\text{max}$)};
\begin{scope}
\filldraw[
    fill=red,
    fill opacity=0.15,
    draw=red,
    dashed,
    thick,
    rounded corners=2pt
]
(B-1-4.north west) --
(B-1-5.north east) --
(B-5-5.south east) --
(B-5-1.south west) --
(B-4-1.north west) --
(B-4-3.north east) --
(B-1-4.south west) --
cycle;
\end{scope}
\node[below=0.5cm of B-5-3, font=\scriptsize] {Zero Padding};
\end{tikzpicture}
\caption{
    Illustration of the padding strategies in LCMs. \textit{Left:} Gaussian noise 
    padding for time-series where \(V < V_{\max}\) and \(L < L_\text{max}\). 
    \textit{Right:} Zero padding of lagged adjacency tensors when \(V < V_{\max}\) 
    or \(\ell < \ell_{\max}\). Red dashed boxes indicate padded regions. 
    $\mathcal{N}$ denotes $\mathcal{N}(0, 0.01)$ for brevity.
}
\label{fig:input-padding}
\end{figure}

\subsection{Datasets} \label{app:datasets}

\subsubsection{Synthetic} \label{app:synthetic}

In line with the synthetic temporal data generators provided in \texttt{Tigramite} \cite{runge2023causal}, we implement a temporal SCM-based time-series generator to produce synthetic datasets for training and validating our LCMs. The generator produces both observational samples by simulating temporal structural equation models with additive noise, as defined in Section \ref{sec:introduction}. Synthetic data allow us to evaluate and compare algorithms under controlled conditions while ensuring reproducibility and flexibility.

The data generation process consists of several steps. Initially, a lagged temporal causal graph is constructed by sampling a random directed acyclic graph (DAG) following the Erd\H{o}s-R\'enyi scheme \cite{brouillard2020differentiable}. Similar to \cite{lawrence2020cdml}, our implementation offers extensive configurability, including the number of variables and samples, the minimum and maximum lags \( \ell_\text{min} \) and \( \ell_\text{max} \), as well as the graph density. To model a variety of realistic causal mechanisms, we incorporate a diverse set of (equiprobable) functional dependencies in the SCM, including both linear and nonlinear relationships, and additive Gaussian noise; examples are summarized in Table~\ref{tab:functions}. Observational samples are generated via ancestral sampling, similar to the procedure used in latent variable models \cite[Section~4.2.5]{murphy2023probabilistic}, by traversing the topological order of the lagged causal graph. The overall sampling procedure is summarized in Algorithm~\ref{alg:scm-ancestral-sampling}. For ensuring stability of the generated time-series samples, we employ a warm-up period as in \cite{runge2023causal}, discarding an initial set of samples before collecting data.

\begin{table}[hbp]
\centering
\caption{Synthetic time-series dataset collections. L = linear, NL = non-linear functional relationships.}
\begin{tabular}{|c|c|c|c|c|c|}
\hline
\textbf{Collection} & \textbf{Pairs} & \textbf{Variables} & \textbf{Timesteps} & \(\ell_{\max}\) & \textbf{Func.\ Rels.} \\
\hline
\texttt{Synthetic\_1} & 100k & \(3\text{--}5\)  & 500 & 3 & L, NL \\
\texttt{Synthetic\_2} & 270k & \(3\text{--}12\) & 500 & 3 & L, NL \\
\texttt{CDML}         & 240  & \(4\text{--}11\) & 500 & 3 & L, NL \\
\hline
\end{tabular}
\label{tab:synthetic-datasets}
\end{table}

An important consideration in synthetic data generation is the stationarity of the resulting time-series. The generator in \texttt{Tigramite} addresses this by performing eigenvalue tests on the stability matrix and aborting the generation if instability is detected, without adjusting the parameters. In contrast, \cite{stein2024embracing} handle instability by repeatedly resampling until stationarity is achieved. Both approaches, however, become impractical when generating large, high-dimensional datasets needed for training our LCMs. We mitigate this issue by wrapping the outcomes of unbounded functional dependencies with bounded functions, such as the hyperbolic tangent \( \tanh \) or the sigmoid \( \sigma \). This strategy caps the growth of unbounded functions, ensuring stationarity without altering the underlying causal relationships, and thereby improves the stability and scalability of our synthetic data generation while preserving causal structure.

\begin{table}[h]
\centering
\caption{Functional dependencies used in synthetic SCM generation.}
\label{tab:functions}
\begin{tabular}{lll}
\toprule
\textbf{Category} & \textbf{Example Function} & \textbf{Description} \\
\midrule
Linear & \( f(x) = a x + b \) & Additive linear effect \\
       & \( f(x_1, x_2) = a_1 x_1 + a_2 x_2 + b \) & Linear combination of parents \\
\midrule
Non-linear & \( f(x) = \alpha_nx^n + \ldots \alpha_0x_0 \) & Polynomial \\
           & \( f(x) = e^x \) & Exponential Transformation \\
           & \( f(x) = \sin(x) \) & Sinusoidal \\
           & \( f(x_1, x_2) = x_1 x_2 \) & Multiplicative interaction \\
           & \( f(x) = \log(1 + |x|) \) & Logarithmic transformation \\
\midrule
Bounded & \( f(x) = \tanh(x) \) & Hyperbolic tangent (bounded in \([-1,1]\)) \\
        & \( f(x) = \sigma(x) \) & Sigmoid \( \big(\frac{1}{1+e^{-x}}\big) \) \\
\bottomrule
\end{tabular}
\end{table}

\subsubsection{Semi-synthetic} \label{app:semi-synthetic}

\begin{table}[!t]
\centering
\small
\caption{Overview of semi-synthetic time-series dataset collections.}
\vspace{5pt}
\begin{tabular}{|c|c|c|c|c|c|}
\hline
\textbf{Collection} & \textbf{Datasets} & \textbf{Vars} & \textbf{Timesteps} & \textbf{Max Lag} & \textbf{Rels.} \\
\hline
\(f\)MRI5 & 21 & 5 & 1200--5000 & 1 & Non-linear \\
\(f\)MRI & 26 & 5--10 & 1200--5000 & 1 & Non-linear \\
\(\text{Kuramoto}\_5\) & 1000 & 5 & 500 & 1 & Non-linear \\
\(\text{Kuramoto}\_10\) & 1000 & 10 & 500 & 1 & Non-linear \\
\hline
\end{tabular}
\label{tab:semi-synthetic-datasets}
\end{table}

Semi-synthetic datasets combine analytically defined structural equations with generative processes and experimental knowledge (e.g., Markov processes of physical systems). These datasets simulate complex systems where the underlying causal graph is known, allowing for controlled benchmarking under quasi-realistic dynamics. An overview is provided in Table \ref{tab:semi-synthetic-datasets}.

The \text{\(f\)MRI} collection comprises 27 datasets with variable counts of 5, 10, and 15, and observation lengths ranging from 50 to 5000, simulating BOLD (Blood-Oxygen-Level Dependent) responses from different regions of interest in the brain. The simulation process follows a non-linear balloon model \cite{buxton1998dynamics}, translating neural activity into hemodynamic responses, and is designed to mimic realistic neural connectivity patterns based on known structural priors. Due to our model input limitations, the 15-variable case is discarded.

\subsubsection{Realistic} \label{app:realistic}

\begin{table}[t!]
\centering
\caption{Overview of real time-series datasets used for generating realistic data pairs, employing the methodology by \cite{gkorgkolis2026adversarial}.}
\vspace{10pt}
\begin{tabular}{|c|c|c|c|}
\hline
\textbf{Collection} & \textbf{Variables} & \textbf{Timesteps} & \textbf{Domain} \\
\hline
WTH & 12 & 35064 & Weather \\
\hline
ETTh1 & 7 & 17420 & Power \\
\hline
ETTm1 & 7 & 69680 & Power \\
\hline
AirQualityUCI & 12 & 9357 & Weather \\
\hline
Bike-usage & 5 & 52584 & Transportation \\
\hline
Outdoors & 3 & 1440 & Environmental \\
\hline
ETTm2 & 7 & 69680 & Power \\
\hline
Climate & 4 & 1463 & Environmental \\
\hline
Garments & 7 & 692 & Manufacturing \\
\hline
power-consumption & 8 & 52416 & Power \\
\hline
Gearbox-fault & 3 & 2000 & Automotive \\
\hline
AirQualityMS & 36 & 4937 & Weather \\
\hline
\end{tabular}%
\label{tab:real-datasets}
\end{table}

To generate realistic (simulated) time-series datasets paired with ground-truth temporal causal graphs, we employ the methodology of \cite{gkorgkolis2026adversarial}, which takes as input real multivariate time-series data, learns a TSCM (the directed causal graph, functional mechanisms, and noise models), and synthesizes arbitrary numbers of time-series samples from the learned causal model. This procedure produces fully supervised dataset--graph pairs suitable for training LCMs.

For any of these input time-series, we verify for stationarity using the \textit{Augmented Dickey-Fuller test} \cite{dickey1979distribution} (ADF). In case non-stationarity is observed, it is normalized using finite differences up to second order. This step assures that data fed into training of LCMs do not possess extreme trends, possibly resulting in unstable training. We follow the python implementation of \texttt{statsmodels} with maximum lag order \( 12 \cdot \left(\frac{T}{100}\right)^{1/4} \) where \(T\) is the number of observed timesteps.

\subsubsection{Used configurations \& algorithms}
We implement the same methods for CD, functional dependency and noise estimation for the generation of simulated data as in \cite{gkorgkolis2026adversarial}\footnote{\url{https://github.com/kougioulis/ACT}}, apart from Random Forest Regressors. We instead use kNN Regressors \cite{fix1985discriminatory} and Gradient Boosting \cite{chen2016xgboost} for improved computational efficiency of generating large amounts of data pairs. An overview is shown in Table \ref{tab:implemented_methods}.

\begin{table}[h!]
\centering
\caption{Implemented Methods for Temporal Functional Dependency Estimation, Noise Density Estimation in Temporal Causal Discovery.}
\setlength{\tabcolsep}{4pt}
\begin{tabular}{ll}
\toprule
\multicolumn{2}{c}{\textbf{Causal Discovery}} \\ 
\midrule
\textbf{Method} & \textbf{Description} \\ 
\midrule
PCMCI \cite{runge2018causal} & Constraint-Based Algorithm \\ 
DYNOTEARS \cite{pamfil2020dynotears} & Constraint Optimization Algorithm \\ 
CP \cite{stein2024embracing} & Pre-trained Transformer \\ 
\midrule
\multicolumn{2}{c}{\textbf{Functional Dependencies}} \\ 
\midrule
Random Forest Regressor \cite{breiman2001random} & Ensemble Learning Regressor \\ 
kNN Regressors \cite{fix1985discriminatory} & Non-parametric Supervised Method \\
XGBoost \cite{chen2016xgboost} & Gradient Boosting Tree Algorithm \\
AD-DSTCN \cite{nauta2019causal} & Time-series Forecasting Model (TCDF module) \\ 
TimesFM \cite{das2024decoder} & Foundational Time-series Forecaster \\ 
\midrule
\multicolumn{2}{c}{\textbf{Noise Density}} \\ 
\midrule
Normal Distribution & Fitted Parametric Distribution \\ 
Uniform Distribution & Fitted Parametric Distribution \\ 
Neural Splines \cite{durkan2019neural} & Normalizing Flows \\ 
RealNVP \cite{dinh2017density} & Normalizing Flows \\ 
\bottomrule
\end{tabular}
\label{tab:implemented_methods}
\end{table}

Apart from the generation of different time-series samples and causal graph data pairs generated from different configurations of algorithms in Phases 1-3 of ACT, we apply two augmentation strategies before processing them through the pipeline to increase the number of simulated datasets: i) time window sub-sampling and ii) node subsets sub-sampling. 

\paragraph{Time Window Sub-sampling} Given a multivariate time-series of minimum length \(L = 400\), we extract multiple non-overlapping windows of fixed length \(w = 200\). Each window is treated as an independent dataset and passed through the full ACT pipeline (causal discovery, function approximation, and noise modeling), resulting in a distinct simulated time-series and corresponding lagged causal graph. 

\paragraph{Node sub-sampling} For each time window, we additionally generate subsets of variables of size \(k \in \{3, 5, 7\}\). Each subset is independently simulated via ACT, with their total count dependent on the dimension of the original dataset. The above strategies increase variability of training pairs by exposing the model to smaller-scale graphs. 

The final collections for training and evaluation are provided in Table \ref{tab:real-dataset-collections}.

\begin{table}[t!]
\centering
\caption{Overview of realistically generated time-series collections for training and OOD evaluation of LCMs. Collections in \textbf{bold} notate inclusion in the \texttt{SIM} training corpus and the large scale, \texttt{LS} corpus.}
\vspace{10pt}
\begin{tabular}{|c|c|c|c|}
\hline
\textbf{Collection} & \textbf{Variables} & \(\mathbf{\ell_{\max}}\) & \textbf{Num. Datasets} \\
\hline
\textbf{WTH} & 3-12 & 3 & 20200 \\
\hline
\textbf{ETTh1} & 3-7 & 3 & 9000 \\
\hline
\textbf{ETTm1} & 3-7 & 3 & 9000 \\
\hline
\textbf{AirQualityUCI} & 3-12 & 3 & 3900 \\
\hline
\textbf{Bike-usage} & 3-5 & 3 & 1800 \\
\hline
\textbf{Outdoors} & 3 & 3 & 800 \\
\hline
ETTm2 & 3-7 & 3 & 15 \\
\hline
Climate & 3,4 & 3 & 17 \\
\hline
Garments & 3-7 & 3 & 24 \\
\hline
Power & 3-8 & 3 & 18 \\
\hline
Gearbox-fault & 3 & 3 & 20 \\
\hline
AirQualityMS & 3-12 & 3 & 54 \\
\hline
\end{tabular}%
\label{tab:real-dataset-collections}
\end{table}

\subsection{Architecture of LCMs} \label{app:lcm-architecture}

This section provides a detailed description of the implemented architecture \cite{zhou2021informer, stein2024embracing} used throughout experiments, consistent with the formulation presented in the main text.

\subsubsection{Input Handling \& Embeddings} \label{subsec:arch-inputs}

\paragraph{Input Embeddings.}
The model receives an observational multivariate time series \(X \in \mathbb{R}^{B \times L \times V}\), where \(B\) denotes the batch dimension, \(L\) the number of timesteps, and \(V\) the number of variables. During inference, \(B=1\). All inputs are min–max normalized per variable. To allow training across datasets with heterogeneous dimensionality and sequence length, inputs are padded (or truncated) to fixed upper bounds \(L_{\max}=500\) and \(V_{\max}=12\), as described in Subsection \ref{subsec:arch-inputs}. Temporal padding is performed using i.i.d.\ Gaussian noise \(z \sim \mathcal{N}(0,0.01)\).

\subsection{Min--Max Normalization} \label{app:minmax}

To ensure consistent scaling across heterogeneous datasets and to stabilize optimization, each variable is min--max normalized prior to being processed. This transformation is applied during both training and inference. For a variable \(x^v_t\), the normalized value \(\tilde{x}^v_t\) is defined as

\begin{equation}
\tilde{x}^v_t
= \frac{x^v_t - \min(x^v)} {\max(x^v) - \min(x^v) + \epsilon},
\end{equation}

where \(\epsilon > 0\) ensures numerical stability.

After min--max normalization, sequences are padded along both temporal and variable dimensions to fixed sizes \(L_{\max}\) and \(V_{\max}\) with sequences longer than \(L_{\max}\) truncated, as in Appendix \ref{app:input-padding}.

\paragraph{Temporal Convolutional Embedding.}

The normalized and padded input is then projected into a latent representation using a one-dimensional convolution operating along the temporal dimension. After permuting the tensor to channel-first format \([B, V_{\max}, L_{\max}]\), a convolution is applied with
\(c_{\text{in}} = V_{\max}, \quad c_{\text{out}} = d_{\text{model}},\) where \(d_{\text{model}}\) denotes the model dimension. The resulting embeddings are \(X_{\text{tok}} = \mathrm{Conv1D}(X) \in \mathbb{R}^{B \times L_{\max} \times d_{\text{model}}}\).

This convolutional projection serves as a learnable feature extractor capturing local temporal dependencies prior to global attention. It replaces the purely linear token embedding of the vanilla Transformer, following the Informer design.

\paragraph{Positional Encoding.}

Because self-attention is permutation-invariant with respect to temporal order, explicit positional information is injected using sinusoidal encodings \cite{vaswani2017attention}:

\begin{equation}
\begin{aligned}
\mathrm{PE}_{(t,2i)} &=
\sin\!\left(
\frac{t}{10000^{2i/d_{\text{model}}}}
\right), \\
\mathrm{PE}_{(t,2i+1)} &=
\cos\!\left(
\frac{t}{10000^{2i/d_{\text{model}}}}
\right).
\end{aligned}
\end{equation}

The final embedding representation is thus \(X_{\text{emb}} = X_{\text{tok}} + \mathrm{PE}\), followed by dropout for regularization.

\subsection{Transformer Encoder}

The embedding representation \(X_{\text{emb}}\) is processed by a stack of Informer-style encoder layers. For a layer \(l\):

\begin{equation}
\begin{aligned}
E'^{(l)} &=
\mathrm{LayerNorm}
\left(
E^{(l)} + \mathrm{SelfAttn}(E^{(l)})
\right), \\
E^{(l+1)} &=
\mathrm{LayerNorm}
\left(
E'^{(l)} + \mathrm{FFN}(E'^{(l)})
\right).
\end{aligned}
\end{equation}

Where \(\mathrm{SelfAttn}(\cdot)\) denotes scaled multi-head self-attention and

\begin{equation}
\mathrm{FFN}(x)
=
\mathrm{Conv1D}_2
\left(
\mathrm{GELU}
\big(
\mathrm{Conv1D}_1(x)
\big)
\right)
\end{equation}

is a two-layer convolutional feed-forward network applied position-wise. Residual connections and layer normalization are also applied after both sublayers to ensure stable propagation of gradients.

\paragraph{Temporal Distillation.}

Between encoder blocks, a self-attention distillation layer is inserted (included in all trained models). It performs convolutional downsampling followed by normalization, activation, and max pooling:

\begin{equation}
X_{\text{distil}}
=
\mathrm{MaxPool}
\left(
\mathrm{ELU}
\big(
\mathrm{BatchNorm}
(
\mathrm{Conv1D}(E^{(l)})
)
\big)
\right).
\end{equation}

This reduces the temporal dimension and lowers computational cost while preserving dominant temporal patterns.

Let the final encoder output be \(E_{\text{enc}} \in \mathbb{R}^{B \times L' \times d_{\text{model}}}\) (\(L' \leq L_{\max} \)) where only the final timestep embedding is retained:
\(h = E_{\text{enc}}[:, -1, :] \in \mathbb{R}^{B \times d_{\text{model}}}\).

This design relies on the global receptive field of self-attention: the final token representation can attend to all preceding timesteps and therefore serves as a global summary of the sequence.

\subsubsection{Auxiliary Correlation Injection}

To assist the model in inferring directed dependencies, empirical Pearson cross-correlations are computed for each ordered variable pair \((X^i, X^j)\) and lag \(\tau \in \{1,\dots,\ell_{\max}\}\), which are flattened and concatenated to the learned neural representations of the final encoder block.

\[
\rho_{X^i,X^j}(\tau)
=
\frac{\mathrm{Cov}(X^i_{1:T-\tau}, X^j_{\tau+1:T})}
{\sqrt{\mathrm{Var}(X^i_{1:T-\tau}) \, \mathrm{Var}(X^j_{\tau+1:T})}}.
\]

These values form a tensor \(\tilde{CC} \in \mathbb{R}^{B \times V_{\max} \times V_{\max} \times \ell_{\max}} \), which is flattened and concatenated with the encoder representation: \(Z = \mathrm{Concat}\!\big( \mathrm{Flatten}(\bar{X}), \mathrm{Flatten}(\tilde{CC})\big)\). This correlation injection acts as a statistical training aid, aligning learned representations with empirically observable lagged dependencies.

\subsubsection{Feedforward Head and Output Interpretation}

The concatenated representation \(Z\) is passed through a two-layer feedforward head with layer normalization and SwiGLU activation: \(H = \mathrm{SwiGLU}\!\big( \mathrm{LN}(W_1 Z + b_1) \big)\), \(\hat{\mathbb{A}}_{\text{flat}} = W_2 H + b_2\).

After reshaping and a sigmoid activation for the transformation of logits, the lagged adjacency tensor corresponding to the discovered lagged causal graph is 

\[
\hat{\mathbb{A}} = \sigma\!\left( \mathrm{Reshape}(\hat{\mathbb{A}}_{\text{flat}}) \right) \in [0,1]^{B \times V_{\max} \times V_{\max} \times \ell_{\max}}\]

Each entry \(\hat{\mathbb{A}}_{j,i,\tau}\) represents the confidence on the directed lagged edge \(X^i_{t-\tau} \rightarrow X^j_t\) existence. Since multiple edges and lags may simultaneously be active, sigmoid activation is used instead of softmax, thus framing temporal causal discovery as a multi-label classification problem.

\paragraph{Limitations.} The flattened output head does not enforce permutation equivariance by design. Future directions can explore equivariant neural models that meet this desiderata by design, while retaining expressivity.

\subsection{Model Sizes \& Training Setup} \label{app:model-sizes}

\begin{table}[ht!]
\centering
\caption{Model configurations and parameter sizes for the LCM variants used in the ablations and additional results in Section \ref{sec:results}.}
\label{tab:model-sizes}
\small
\begin{tabular}{lcccccc}
\toprule
\textbf{Model} & \textbf{\#Params} & \textbf{Enc. Layers} & \(d_\text{model}\) & \textbf{Attn. Heads} & \(d_\text{ff}\) & \textbf{Usage} \\
\midrule
- & \texttt{905K-914K}   & 1  & 256 & 2 & 128 & Subsection \ref{subsec:exp-correlations} \\
- & \texttt{1M}  & 1 & 256 & 2 & 256 & Subsection \ref{subsec:exp-synth-sim} \\
LCM-\texttt{2.5M}  & \texttt{2.5M}   & 4  & 256 & 4 & 256 & Subsection \ref{subsec:res-lcms} \\
LCM-\texttt{9.4M}  & \texttt{9.4M}   & 4  & 512 & 4 & 512 & Subsection \ref{subsec:res-lcms} \\
LCM-\texttt{12.2M}  & \texttt{12.2M}   & 4  & 512 & 4 & 512 &Subsection \ref{subsec:res-lcms} \\
LCM-\texttt{24M}   & \texttt{24M}  & 8  & 1024 & 8 & 1024 & Subsection \ref{subsec:res-lcms} \\
\bottomrule
\end{tabular}
\end{table}

Table \ref{tab:model-sizes} summarizes the implemented models and their configurations used in Section \ref{sec:results}. 

All models are implemented in Python~3.10 using PyTorch~2.2 \cite{paszke2019pytorch}. Training uses an effective batch size of 64, with gradient accumulation when necessary to accommodate larger models, using the AdamW optimizer \cite{loshchilov2019adamw} and a learning rate of \(1e-4\) throughout. Learning rate scheduling and early stopping based on validation AUC are employed to ensure stable convergence. Training was conducted on a dedicated workstation equipped with a NVIDIA RTX 4090 GPU (32 GB VRAM) and 128 GB physical memory. Training of large-scale models took around two weeks to complete. Causal Discovery (LCM Inference step) is performed on an AMD Ryzen 5 CPU. 

\subsection{Model Evaluation Methodology} \label{app:baselines}

\begin{algorithm}[]
\caption{Bootstrap-based Estimation of Edge Probabilities} \label{alg:bootstrap}
\begin{algorithmic}[1]
\REQUIRE Time-series dataset $\mathcal{D}$, maximum lag $\ell_{\max}$, number of bootstraps $n$, causal discovery algorithm configuration $\mathbf{B}_\text{CD}$
\ENSURE Soft adjacency matrix $\mathbb{A} \in [0,1]^{V \times V \times \ell_{\max}}$

\STATE Initialize accumulator $\mathbf{S} \gets \mathbf{0}^{V \times V \times \ell_{\max}}$

\FOR{$b = 1$ to $n$}
    \STATE \COMMENT{Bootstrap Sampling Phase}
    \STATE Draw bootstrap sample $\mathcal{D}^{(b)}$ from $\mathcal{D}$ by resampling with replacement

    \STATE \COMMENT{Causal Discovery Phase}
    \STATE $\mathbf{A}^{(b)} \gets \text{causalAlg}(\mathcal{D}^{(b)}, \mathbf{B}_\text{CD})$

    \STATE \COMMENT{Edge Confidence Update}
    \STATE $\mathbf{S} \gets \mathbf{S} + f(\mathbf{A}^{(b)})$ 
    \STATE \COMMENT{$f$ extracts edge indicators or confidence scores}
\ENDFOR

\STATE \COMMENT{Normalization Phase}
\STATE $\mathbf{A} \gets \mathbf{S} / n$

\RETURN $\mathbf{A}$
\end{algorithmic}
\end{algorithm}

To ensure a fair and consistent evaluation across all causal discovery methods, we compute performance metrics based on each model’s native output format while adapting our evaluation pipeline accordingly. For models that produce real-valued edge confidences (our LCMs and CausalPretraining models), the values are used verbatim as continuous scores for computing the Area Under the ROC Curve (AUC) and the (binarized) ground-truth adjacency labels. For constraint-based methods like PCMCI, we use the inverse of the edge p-values as confidence scores. For models that output an adjacency matrix of causal effect coefficients (like VARLiNGAM and DYNOTEARS), results are not directly comparable to our LCMs that output soft adjacency probabilities and computing AUCs is not directly applicable. For a fair comparison, we follow a bootstrap-based procedure to estimate the probability of each causal edge, as shown in the official documentation\footnote{\url{https://lingam.readthedocs.io/en/latest/tutorial/var.html}}. Specifically, we run VARLiNGAM \( n=10\) times with resampled datasets and compute the proportion of times each edge is discovered, resulting in a soft adjacency matrix of edge probabilities. We follow the same procedure for DYNOTEARS and the same number of boostrapped datasets, as in Algorithm \ref{alg:bootstrap}.

\subsection{Pseudocodes \label{app:algorithms}}

\begin{algorithm}[ht!]
\caption{Temporal SCM Ancestral Sampling (ANCESTRAL)}
\label{alg:scm-ancestral-sampling}
\KwIn{Temporal Causal Graph \(\mathcal{G}\) of time-series \(\{\mathbf{X_t}\}_{t=1}^{\ell_\text{max}}\), causal discovery algorithm configuration \(\mathbf{B}_\text{CD}\), predictive model configuration \(\mathbf{B}_\text{pred}\), noise estimator configuration \(\mathbf{B}_\text{noise}\), max lag \(\ell_{\text{max}} \in \mathbf{B}_\text{CD}\), number of timesteps \(T\), number of warmup steps \(W\)}
\KwOut{Generated time-series sample \(\{\mathbf{X}_t\}_{t=1}^{T}\)}
\BlankLine

\textbf{Initialize} \(X^i_{t=0}\) for all \(i \in \{1, \dots, N\}\) with random noise\;

\tcp{Forward Sampling}
\For{\(t = W + \ell_{\text{max}} + 1\) \KwTo \(T\)}{
    \For{each variable \(X^i_t\) in topological order based on \(\mathcal{G}\)}{
        Determine lagged parents \(\text{Pa}_{X^i_t} \gets \{X^j_{t-k} \mid (X^j, X^i) \in \mathcal{G}, 1 \leq k \leq \ell_{\text{max}} \}\) \tcp*{Lagged parents \(X^j\) of \(X^i\) at time \(t\)}
        Compute \(X^i_t \coloneqq f_i(\text{Pa}^i_t) + \epsilon^i_t\)\;
    }
}
\textbf{Return} generated time-series dataset \(\{ \mathbf{X}^i_t\}_{W+\ell_\text{max}}^{T}\)
\end{algorithm}

Algorithm \ref{alg:scm-ancestral-sampling} describes the ancestral sampling procedure for obtaining multivariate time-series samples from a known TSCM. Initial warming steps are discarded as a way to ensure stability \cite{runge2018causal}.

\subsection{Illustrative Example \& Visualizations} \label{app:example}

To provide an intuitive demonstration of LCMs, we construct a synthetic temporal causal process consisting of three variables \(V^1, V^2\) and \(V_3\). The example is governed by a simple, linear temporal structural causal model:
\[
V^1_t := \epsilon_1(t), \quad 
V^2_t := 3 V^1_{t-1} + \epsilon^2_t, \quad 
V^3_t := V^2_{t-2} + 5 V^1_{t-3} + \epsilon^3_t,
\]

where \(\epsilon^i_t \sim \mathcal{N}(0,1) ~ \forall i \in \{1,2,3\}, t \in \mathbb{Z}\). The TSCM thus induces the following lagged causal relationships:

\[
V^1 \xrightarrow[\text{lag}=1]{} V^2, \quad 
V^1 \xrightarrow[\text{lag}=3]{} V^3, \quad 
V^2 \xrightarrow[\text{lag}=2]{} V^3.
\]

\begin{figure}[t!]
\centering
\includegraphics[width=\textwidth]{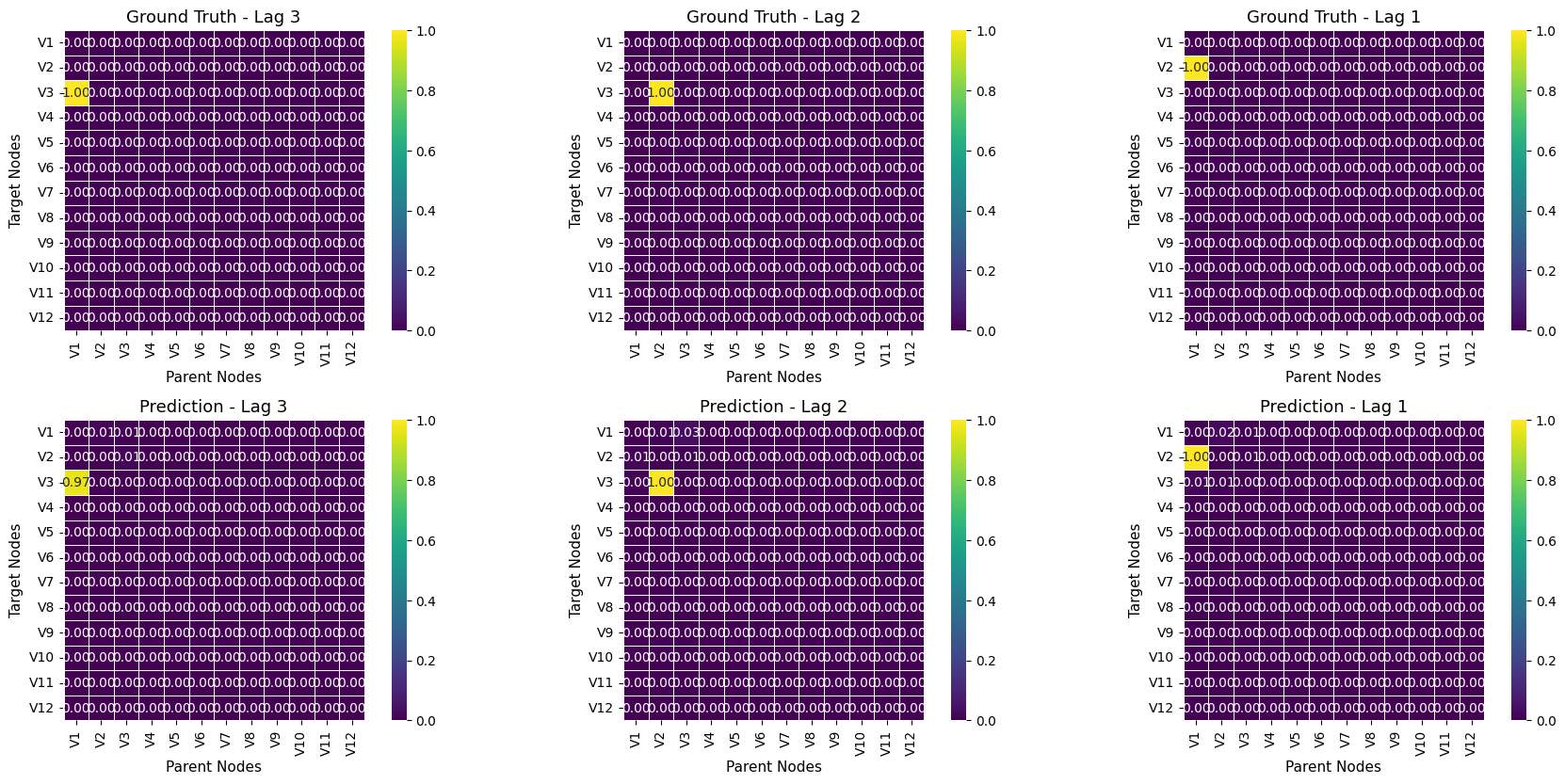}
\caption{Heatmap visualization of the discovered lagged adjacency matrix (bottom row) compared to the ground truth (top row) in an example with \(n_\text{max}=12\) and \(\ell_\text{max}=3\) using the pretrained \texttt{LCM-2.4M} model. Brighter colors in the predicted adjacency indicate stronger confidence for edge existence.} \label{fig:heatmap-example}
\end{figure}

In Figure \ref{fig:heatmap-example} we visualize the discovered causal graph of the final feedforward head against the ground truth; the pretrained LCM successfully captures the underlying causal structure of this simple temporal dataset, resulting in a perfect AUC score of \( 1.00\).

\subsection{Statistical Significance of Results} \label{app:auc-stat-tests}

We evaluate statistical significance of AUC differences between LCMs and baseline methods using the Wilcoxon signed-rank test \cite{conover1999practical}, a non-parametric paired test appropriate for potentially non-Gaussian AUC distributions. To control the family-wise error rate across multiple pairwise comparisons, we apply Bonferroni correction (\(\alpha_{\text{corrected}} = \alpha/k\)) where \(k\) is the number of performed comparisons. 

For each dataset collection, we report the median paired difference in AUC (\(\Delta_{\text{AUC}}\)) relative to a common reference model (the highest average AUC per collection). 

\begin{sidewaysfigure}
\centering
\includegraphics[width=1\textheight]{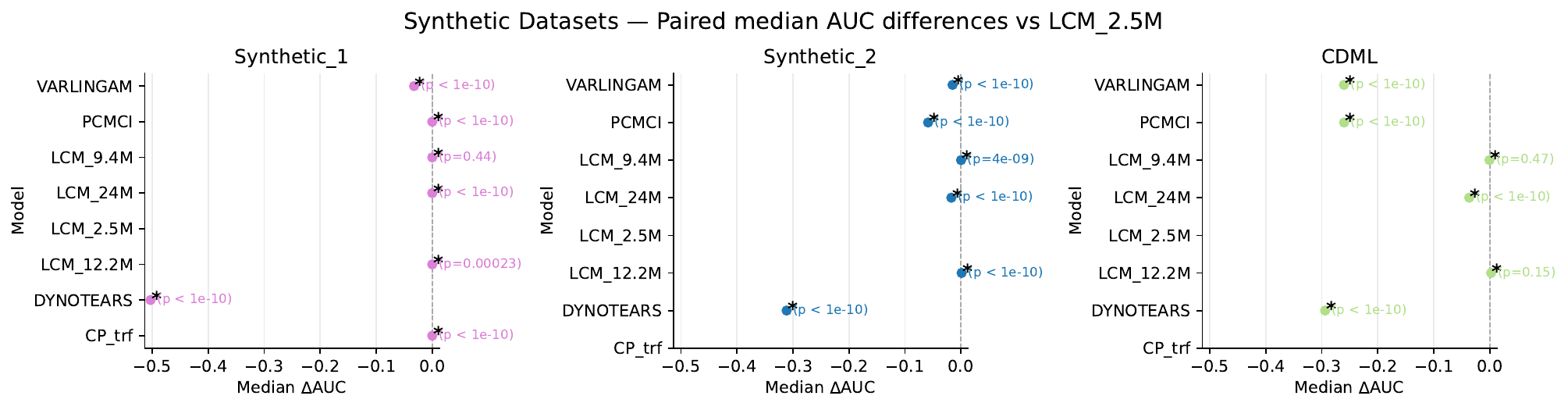}
\caption{Paired median AUC differences relative to the reference model across \textit{Synthetic} collections. Positive values indicate improved performance. Statistical significance is determined via Wilcoxon signed-rank test with Bonferroni correction.}
\label{fig:wilcoxon-synthetic}
\end{sidewaysfigure}

\begin{sidewaysfigure}
\centering
\includegraphics[width=0.7\textheight]{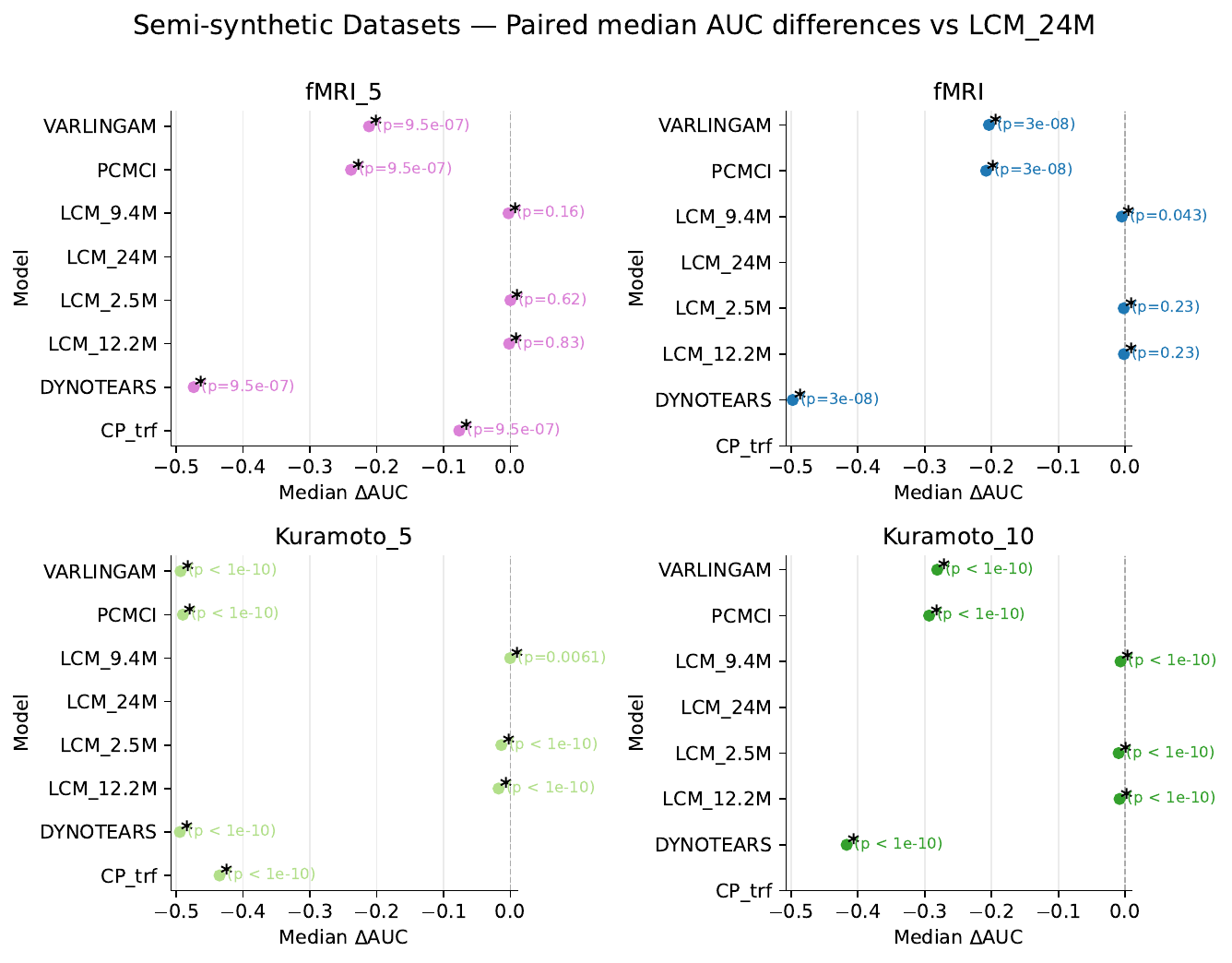}
\caption{Paired median AUC differences across \textit{Semi-Synthetic} collections. Positive values indicate improved performance. Statistical significance is determined via Wilcoxon signed-rank test with Bonferroni correction.}
\label{fig:wilcoxon-semi-synthetic}
\end{sidewaysfigure}

\begin{sidewaysfigure}
\centering
\includegraphics[width=\textheight]{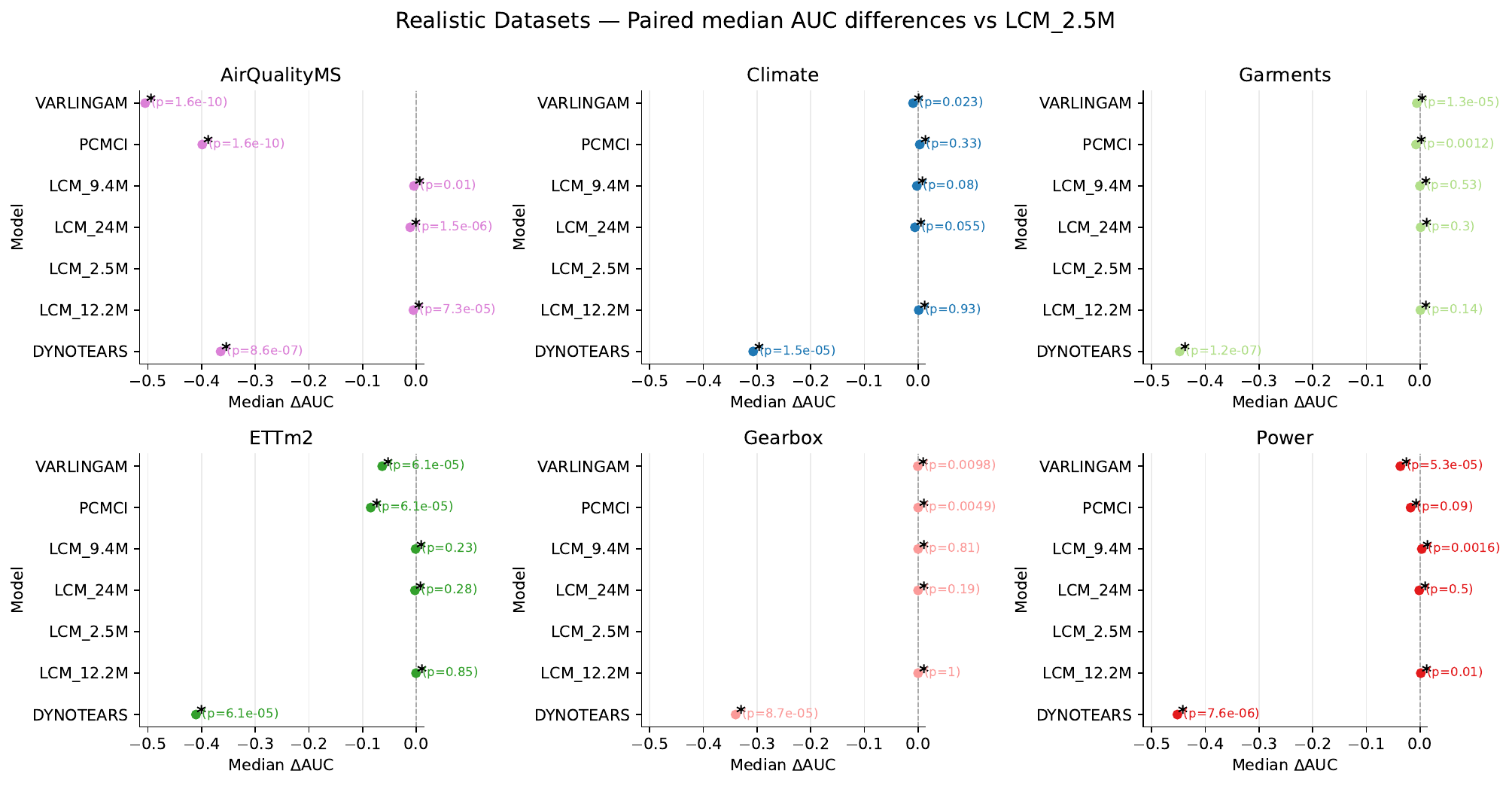}
\caption{Paired median AUC differences across \textit{Realistic} collections. Positive values indicate improved performance. Statistical significance is determined via Wilcoxon signed-rank test with Bonferroni correction.}
\label{fig:wilcoxon-realistic}
\end{sidewaysfigure}

Figures~\ref{fig:wilcoxon-synthetic}–\ref{fig:wilcoxon-realistic} visualize the paired effects from Subsection \ref{subsec:res-lcms}. Asterisks denote statistically significant differences after correction.

\end{document}